\documentclass[10pt,twocolumn,letterpaper]{article}

\usepackage{3dv}
\usepackage{times}
\usepackage{epsfig}
\usepackage{graphicx}
\usepackage{amsmath}
\usepackage{amssymb}
\usepackage{subcaption}
\usepackage{verbatim}
\usepackage{algorithmic}
\usepackage{textcomp}
\usepackage{multirow}
\usepackage[export]{adjustbox}

\usepackage{enumitem}

\usepackage{amsfonts}
\usepackage{color}
\usepackage{listings}
\usepackage{caption}

\lstnewenvironment{algorithm}[1][] %
{   
    \lstset{ %
        mathescape=true,
        frame=tB,
        numberstyle=\tiny,
        basicstyle=\scriptsize, 
        keywordstyle=\color{black}\bfseries\em,
        keywords={,input, output, return, datatype, function, in, if, else, foreach, while, begin, end, for, elif, repeat, times, and,} %
        numbers=left,
        xleftmargin=.06\textwidth,
        #1 %
    }
}
{}

\makeatletter
\renewcommand\paragraph{\@startsection{paragraph}{4}{\z@}%
    {0.9ex \@plus1ex \@minus.3ex}%
    {-1em}%
    {\normalfont\normalsize\bfseries}}
\makeatother

\newcommand{\cell}	{c}
\newcommand{\Cells}	{\mathcal{C}}

\newcommand{\face}	{f}
\newcommand{\Faces}	{\mathcal{F}}

\newcommand{\lineseg}	{l}
\newcommand{\Linesegs}	{\mathcal{L}}
\newcommand{\Linesegsof}{\Lambda}

\newcommand{\Planesof}	{\Pi}

\newcommand{\viewpoint}	{v}
\newcommand{\Viewpoints}{\mathcal{V}}

\newcommand{\plane}	{P}
\newcommand{\altplane}	{\plane'}
\newcommand{\Planes}	{\Pi}
\newcommand{\occup}	{x}
\newcommand{\voccup}	{\mathbf{x}}

\newcommand{\abs}[1]	{|\,#1\,|}
\newcommand{\card}[1]	{|\,#1\,|}
\newcommand{\len}[1]	{|\,#1\,|}

\newcommand{\ifont}[1]{{\textsf{#1}}}

\newcommand{\iprim}	{\ifont{prim}}

\newcommand{\ivis}	{\ifont{vis}}
\newcommand{\iregul}	{\ifont{regul}}

\newcommand{\iedge}	{\ifont{edge}}
\newcommand{\icorner}	{\ifont{corner}}

\newcommand{\ibest}	{\ifont{best}}
\newcommand{\iiter}	{\ifont{iter}}
\newcommand{\imax}	{\ifont{max}}
\newcommand{\ifus}	{\ifont{fus}}

\def\clap#1{\hbox to 0pt{\hss#1\hss}}
\def\mathclap{\mathpalette\mathclapinternal}
\def\mathclapinternal#1#2{%
           \clap{$\mathsurround=0pt#1{#2}$}}

\newcommand{\inhouse}{HouseInterior}

\usepackage[pagebackref=true,breaklinks=true,letterpaper=true,colorlinks,bookmarks=false]{hyperref}

\threedvfinalcopy %

\def\threedvPaperID{207} %

\ifthreedvfinal\fi
\begin{document}

\title{Surface Reconstruction from 3D Line Segments}

\author{Pierre-Alain Langlois\\
 Laboratoire d'informatique Gaspard Monge\\
 Ecole des Ponts ParisTech\\
 Champs-sur-Marne, France
{\tt\small pierre-alain.langlois@enpc.fr}
\and
Alexandre Boulch\\
DTIS, ONERA, University Paris Saclay,\\ F-91123 Palaiseau, France\\
{\tt\small alexandre.boulch@onera.fr}
\and
Renaud Marlet\\
 Laboratoire d'informatique Gaspard Monge\\
 Ecole des Ponts ParisTech\\
 Champs-sur-Marne, France
{\tt\small renaud.marlet@enpc.fr}
}

\author{Pierre-Alain Langlois$^1$ \quad Alexandre Boulch$^2$ \quad Renaud Marlet$^{1,3}$ \\ \small
$^1$LIGM (UMR 8049), ENPC, UPE, France \qquad $^2$ONERA, Universit\'e Paris-Saclay, Palaiseau, France \qquad $^3$valeo.ai, Paris, France}

\maketitle

\begin{abstract}
   In man-made environments such as indoor scenes, when point-based 3D reconstruction  fails due to the lack of texture, lines can still be detected and used to support surfaces. We present a novel method for watertight piecewise-planar surface reconstruction from 3D line segments with visibility information. 
   First, planes are extracted by a novel RANSAC approach for line segments that allows multiple shape support. Then, each 3D cell of a plane arrangement is labeled full or empty based on line attachment to planes, visibility and regularization. Experiments show the robustness to sparse input data, noise and outliers.
\end{abstract}

\section{Introduction}
\label{sec:intro}

Numerous applications make use of 3D models of existing objects%
. In particular, models of existing buildings (e.g., BIMs) allow virtual visits and work planning, as well as simulations and optimizations, e.g., for thermal performance, acoustics or lighting.
The building geometry is often reconstructed from 3D point clouds captured with lidars or using cameras and photogrammetry. But with cameras, registration and surface reconstruction often fail on indoor environments because of the lack of texture and strong view points changes: salient points are scarce, point matching is difficult and less reliable, and when calibration nonetheless succeeds, generated points are extremely sparse and reconstructed surfaces suffer from holes and inaccuracies.

Yet, recent results hint it is possible to rely on line segments rather than points. Lines are indeed prevalent in man-made environments, even if textureless.
From robust detection~\cite{ipol.2012.gjmr-lsd,SalaunICPR2016} and matching \cite{ZhangICISP2011,MSLD,LEHF} to camera registration \cite{ElqurshCVPR2011, SalaunECCV16, Salaun3DV2017,MiraldoECCV2018} and 3D segment reconstruction \cite{HoferBMVC2013,Hofer:cviu2016}, lines can be used when photometric approaches fail for lack of texture. But as opposed to point processing, line-based surface reconstruction has little been studied \cite{WittICRA2014,MentgesICRA2016}.
This paper presents a novel approach to do so.%

\begin{figure}[!t]
\centering
\vspace{-1mm}
\begin{tabular}{@{}c@{}c@{}c@{}}
\includegraphics[width=0.333\linewidth]{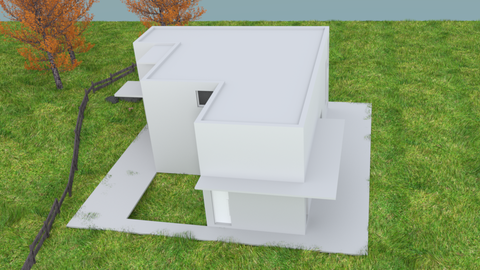} &
\includegraphics[width=0.333\linewidth]{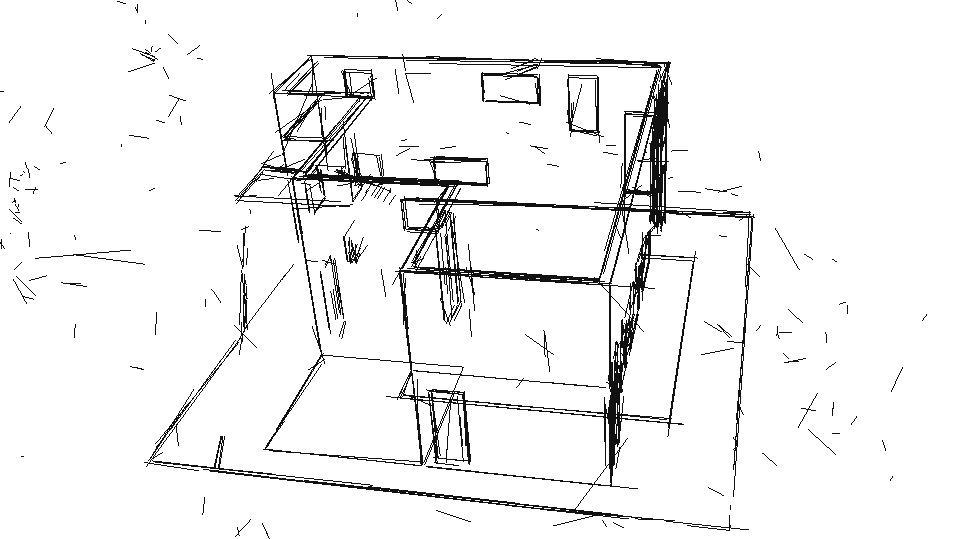} &
\includegraphics[width=0.333\linewidth]{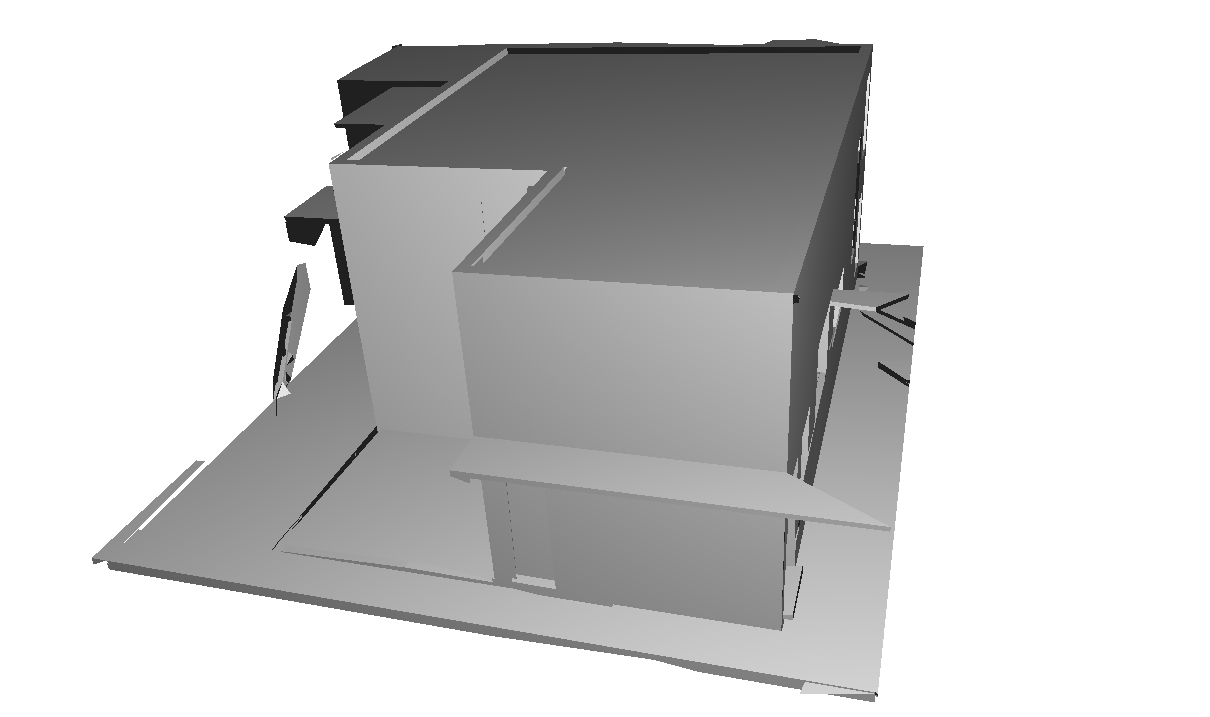}
\\
\includegraphics[width=0.333\linewidth]{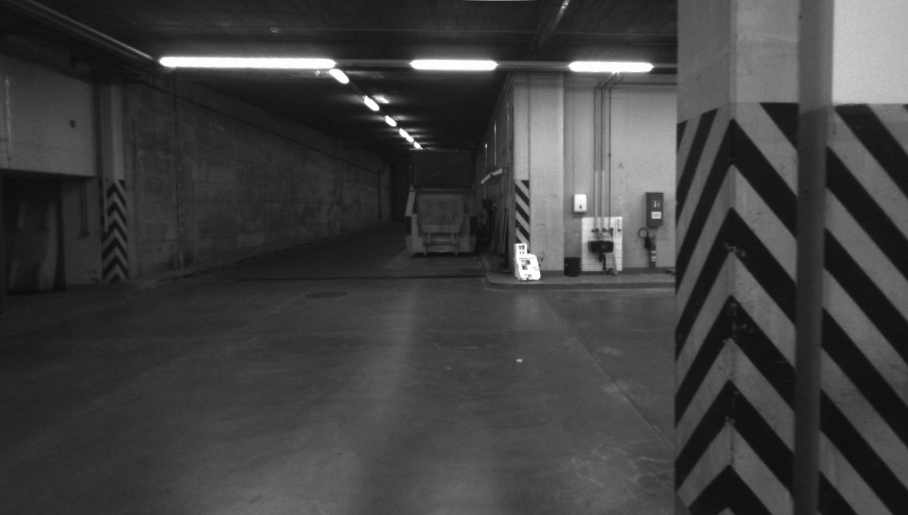} &
\includegraphics[width=0.333\linewidth]{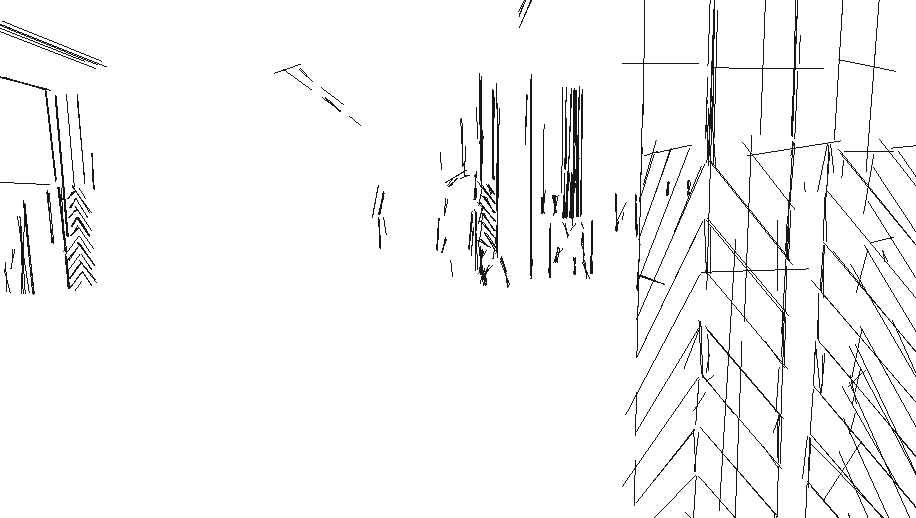} & 
\includegraphics[width=0.333\linewidth]{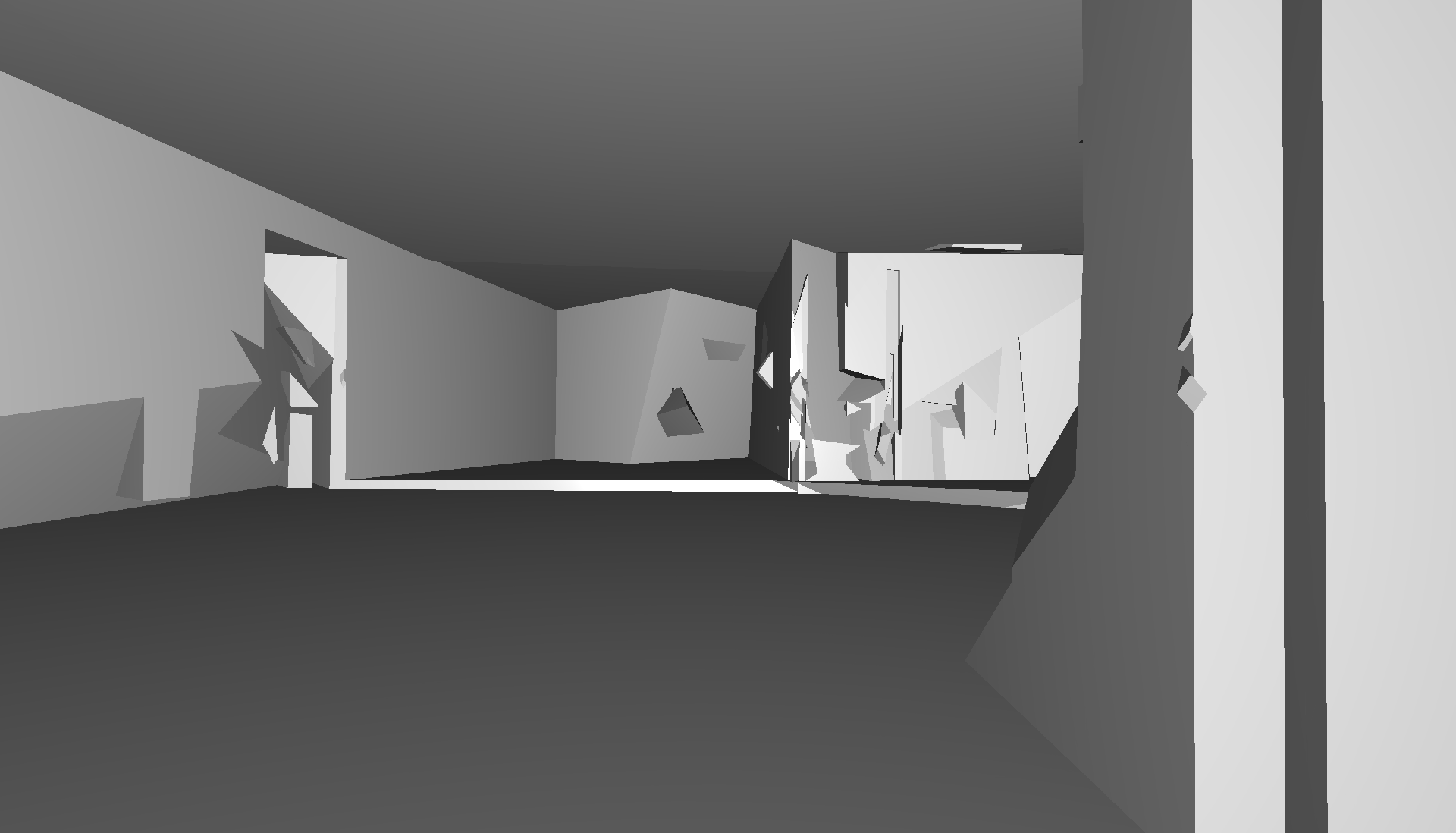}
\\
\includegraphics[width=0.333\linewidth]{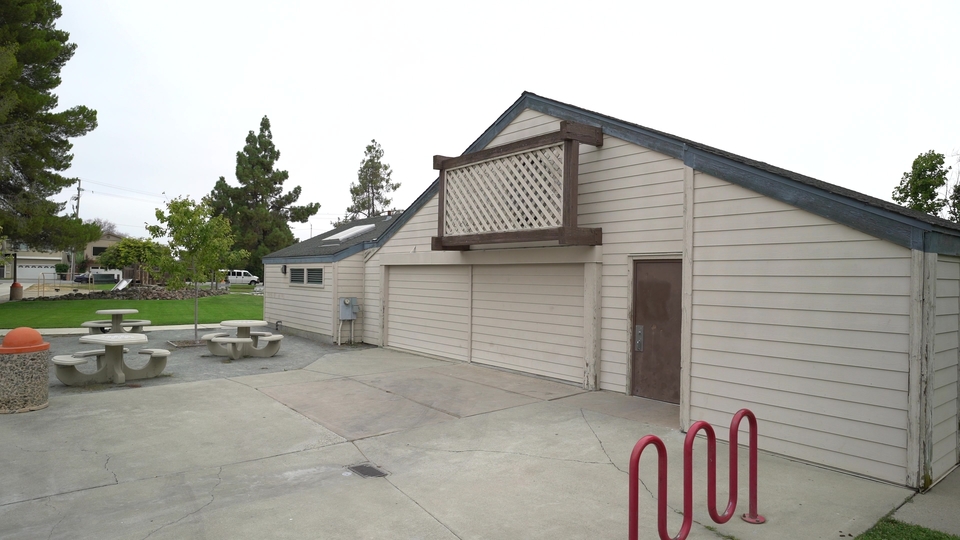} &
\includegraphics[width=0.333\linewidth]{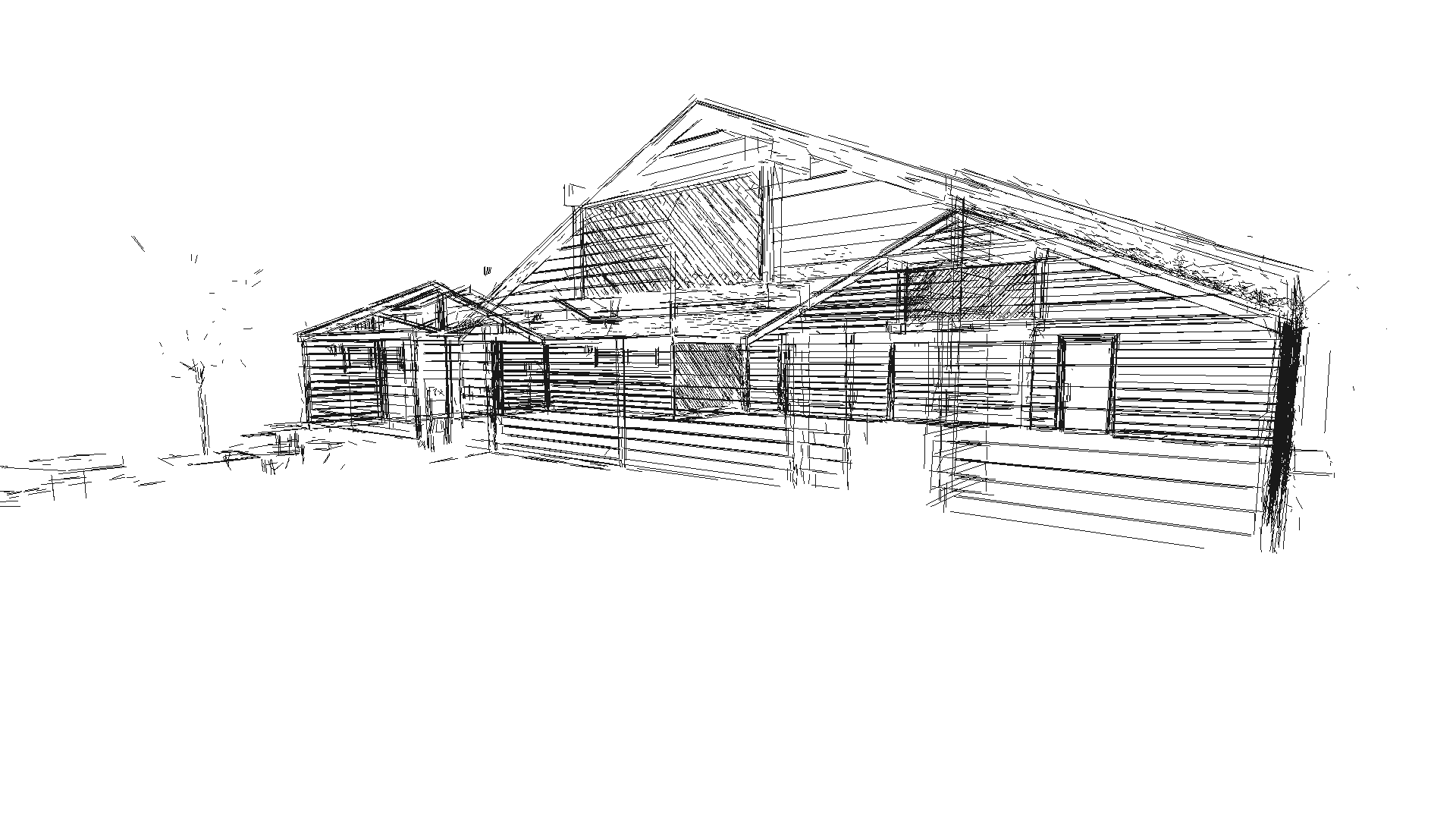} & 
\includegraphics[width=0.333\linewidth]{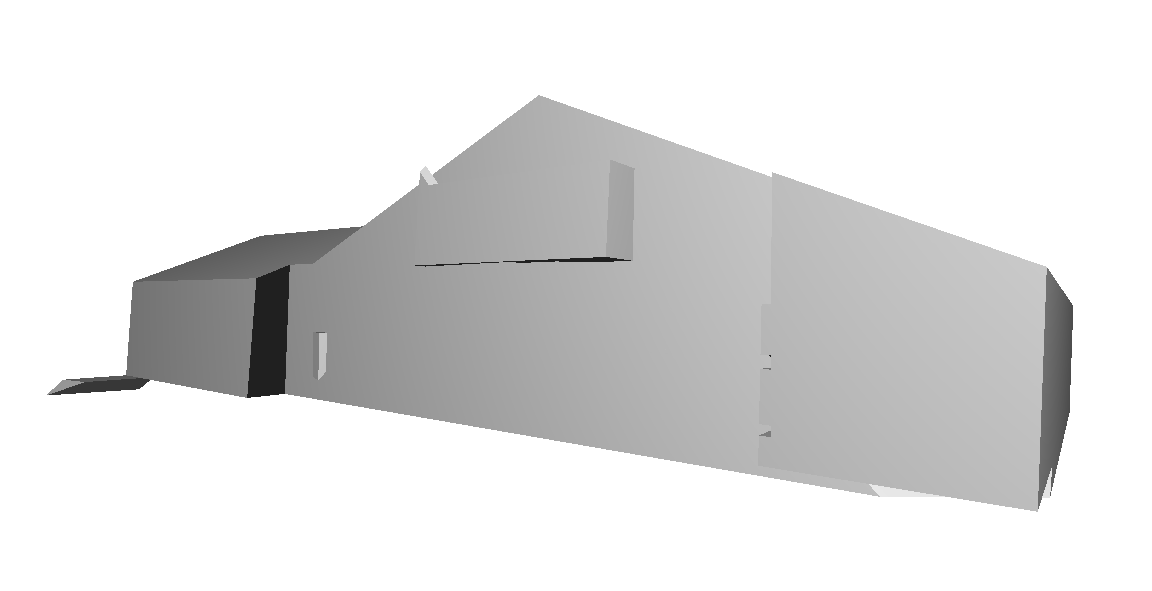}
\\
\includegraphics[width=0.333\linewidth, trim={0 2cm 0 0}, clip]{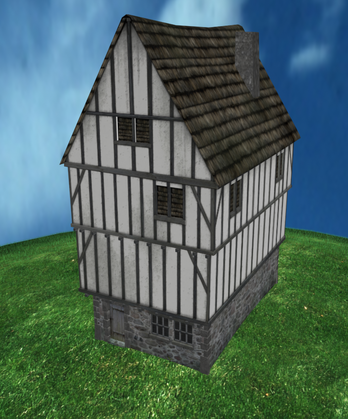} &
\includegraphics[width=0.333\linewidth, trim={0 2cm 0 0}, clip]{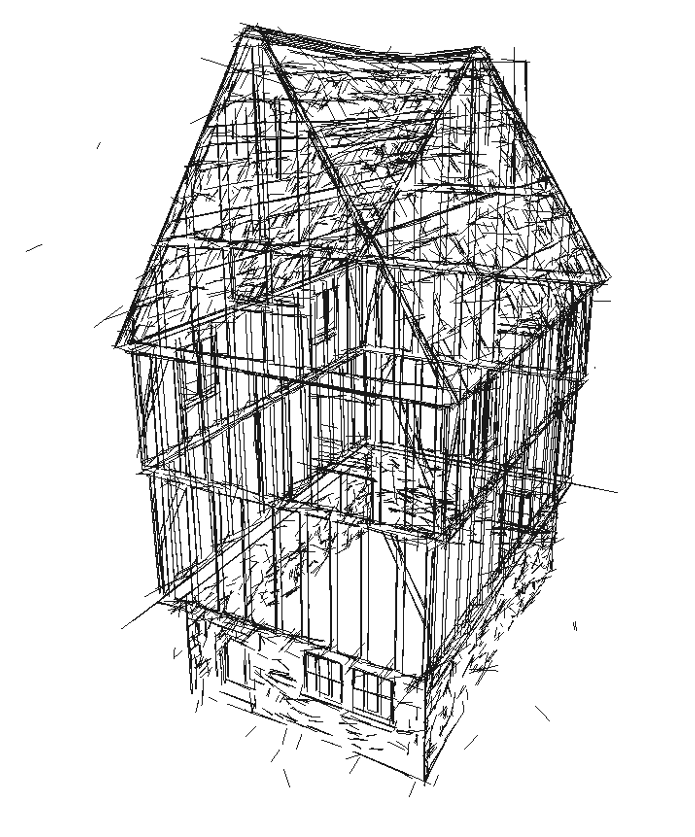} &
\includegraphics[width=0.31\linewidth, trim={0 1cm 0 0}, clip]{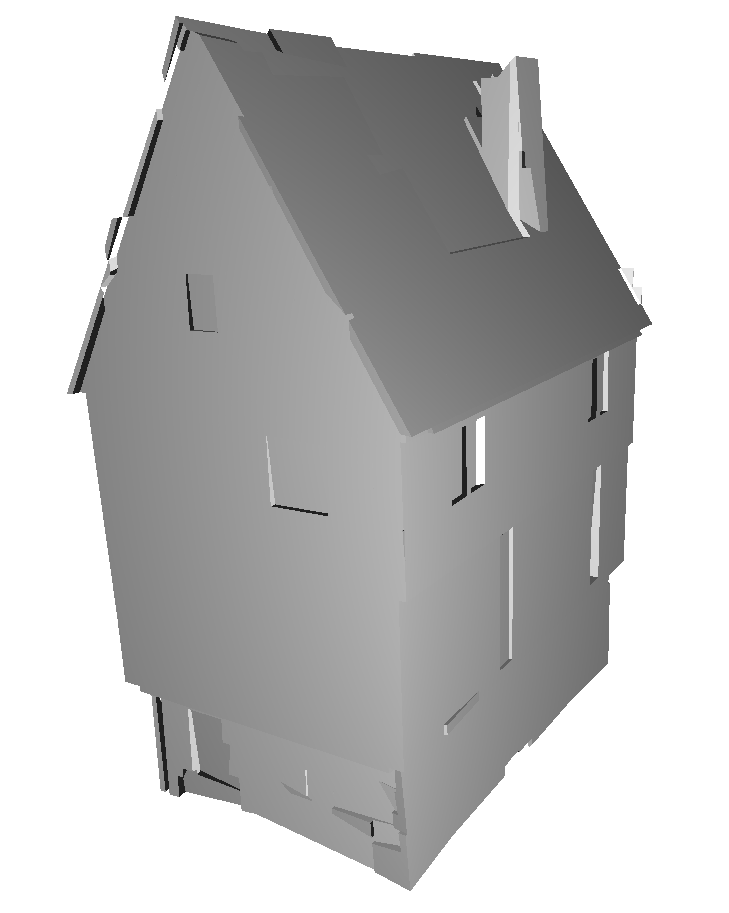}
\\
\includegraphics[width=0.333\linewidth, trim={0 2cm 0 0}, clip]{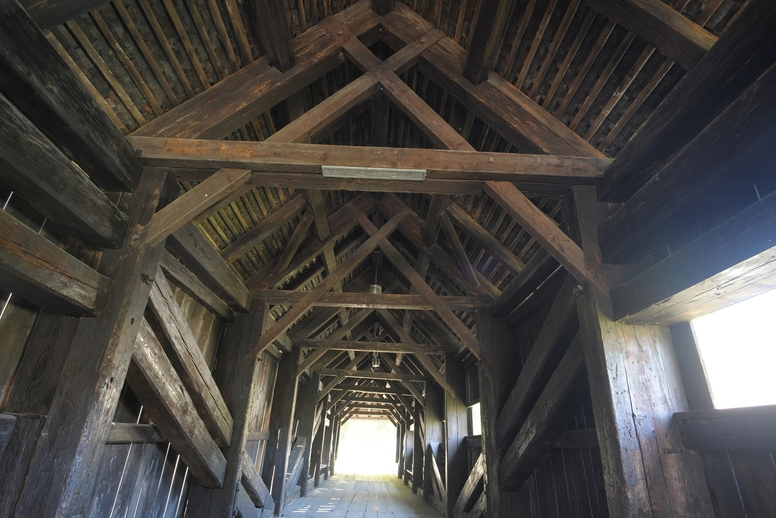} &
\includegraphics[width=0.333\linewidth, trim={0 2cm 0 0}, clip]{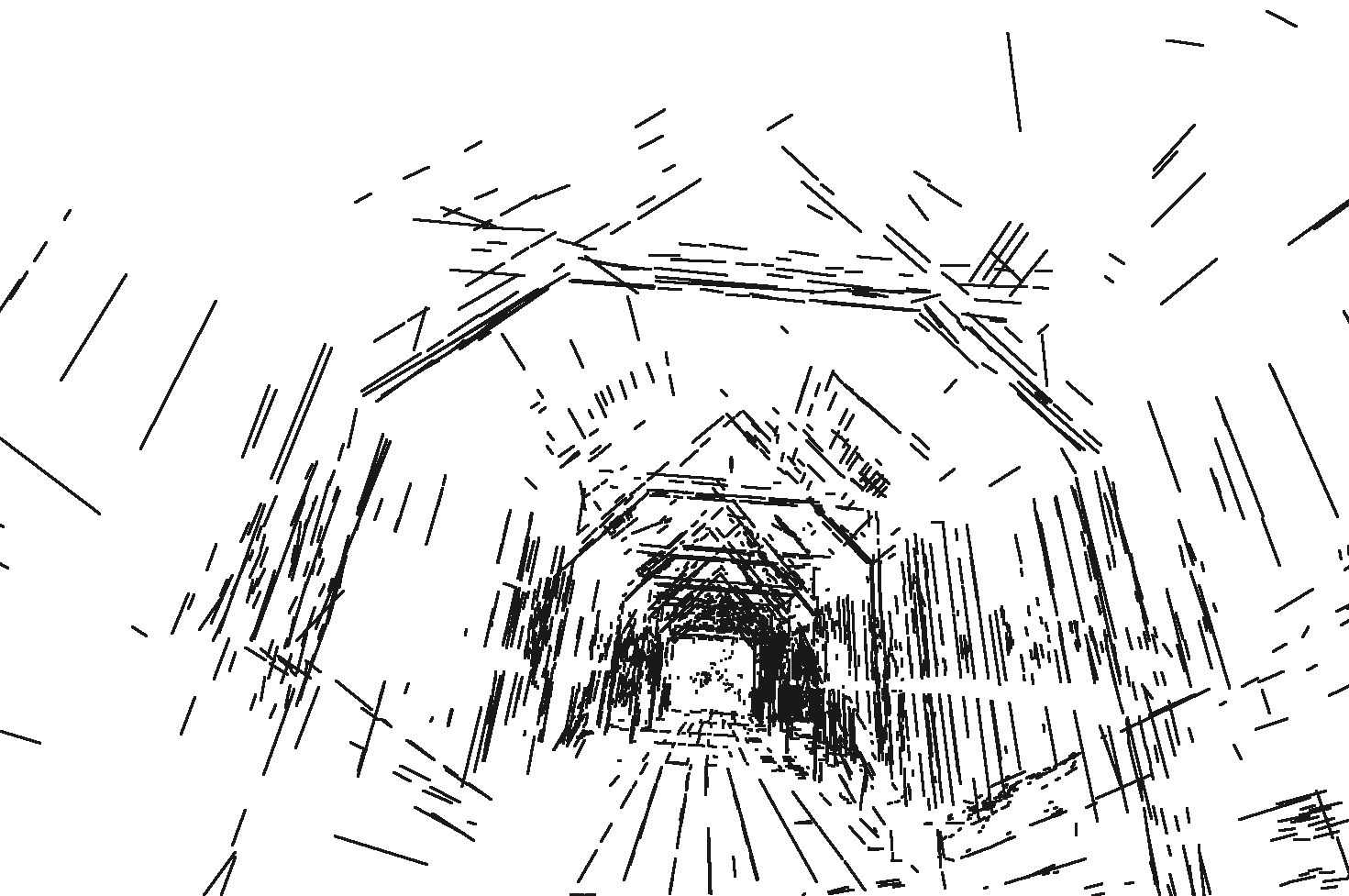} &
\includegraphics[width=0.31\linewidth, trim={0 1cm 0 0}, clip]{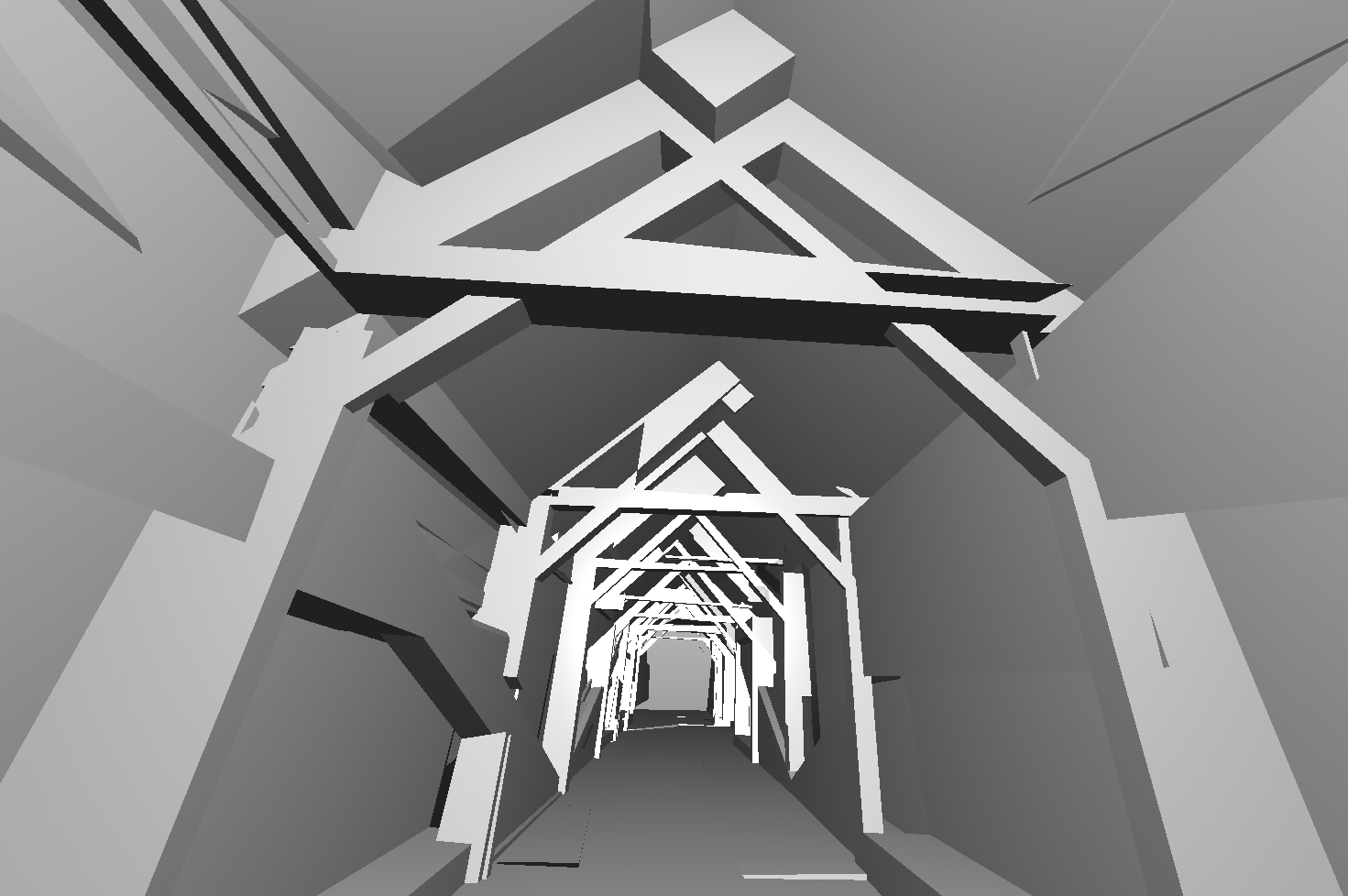}
\end{tabular}
\vspace{-3.5mm}
\caption{Datasets (from top to bottom) Andalusian, DeliveryArea, Barn, Tim\-ber\-Frame, Bridge: (from left to right) image sample, 3D line segments, our reconstruction.}
\vspace{-4.5mm}
\label{fig:datasets}
\end{figure}

\textbf{A change of paradigm} is needed to consider 3D line segments rather than points.
Transposing point-based methods to lines is difficult as many point-related assumptions do not hold for line segments. Indeed, points should be numerous enough (often, in thousands), with a uniform enough sampling, with an accurate enough detection and matching, and most of all, they must belong at most to one primitive.
On the contrary, only a few tens of lines (rarely hundreds) are typically detected, and their density and sampling uniformity is so low that they cannot directly support a good surface reconstruction. %
Also, due to noise in local gradients and varying occlusions depending on viewpoints, segment detection is less accurate and often leads to over-segmentation and unstable end-points, ignored by most 2D line matchers.
Only after image registration and 3D segment reconstruction can 2D detections be related to actual fragments of a 3D line segment, moreover possibly differing according to the different viewpoints.
Besides, curvy shapes as cylinders may yield unstable occlusion edges (silhouettes), yielding noise or outliers.
Finally, some 3D lines identify straight edges that are creases between two planar surfaces, and thus support two shapes, contrary to points.

Belonging to two primitives rather than one requires reconsidering shape detection. In particular, in greedy iterative methods, removing all data supporting a detected shape could prevent detecting other shapes because all or a significant fraction of features would then be missing.  For instance, it would not be possible to detect all the faces of a cube given only its edges.  And even if enough 3D data remains for detection, shape sampling would be affected and some shapes would be less likely or unlikely to be detected.

\begin{figure}[t]
\begin{center}
\includegraphics[width=0.9\linewidth]{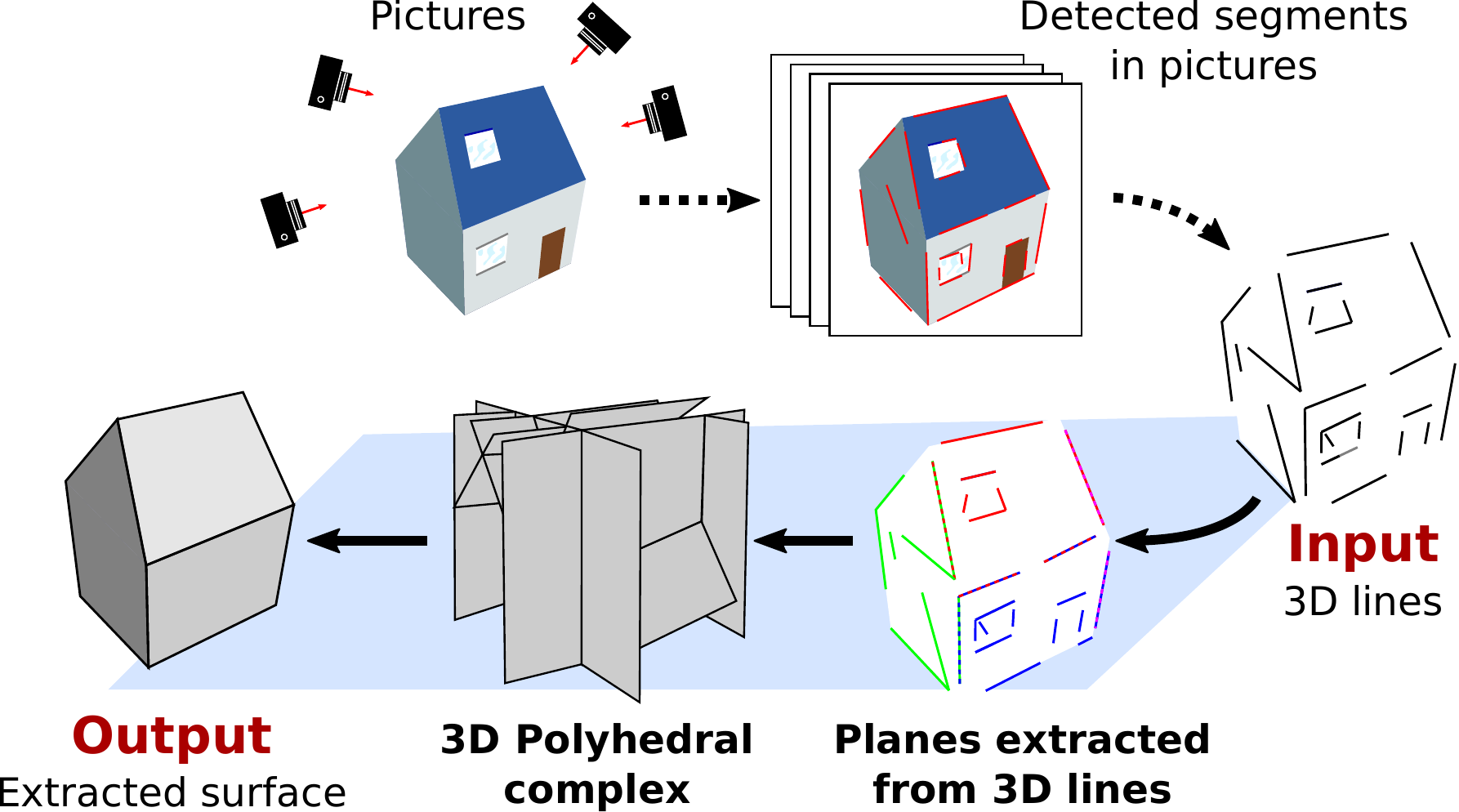}
\end{center}
\vspace*{-4mm}
   \caption{Line-based 3D reconstruction pipeline. This paper covers from \textbf{\small\textsf{Input}} to \textbf{\small\textsf{Output}}.}
\label{fig:pipeline}
\vspace*{-4mm}
\end{figure}

\textbf{Overview.} 
We propose the first complete reconstruction pipeline that inputs 3D line segments with visibility information and outputs a watertight piecewise-planar surface without self-intersection (cf.\ Fig.~\ref{fig:pipeline}).
We first extract primitive planes from the line cloud, distinguishing two kinds of line segments: \emph{textural lines}, supporting a single plane, and \emph{structural lines}, at the edge between two planes. Then we label each 3D cell of the plane arrangement as full or empty by minimizing an energy based on line type, line segment support, visibility constrains and regularization. %

\textbf{Our main contributions} are as follows:\\
\indent - We define a robust and scalable plane detection method from 3D line segments, without scene assumptions. This novel non-straightforward RANSAC formulation takes into account a key specificity of lines vs points, namely that they can support up to two primitives (at edges), which breaks the greedy iterative detection traditionally used with points. \\
\indent - We totally recast the surface reconstruction approach of \cite{Chauve:cvpr2010, Boulch:sgp2014} into a line-based setting.
We meaningfully and efficiently generalize data fidelity and visibility from points to line segments, taking care of lines supporting two planes.
We also feature a simpler and lighter treatment of noise. \\
\indent - We validate our method on existing datasets, and provide new ones to assess line-based reconstruction quality. Examples of our reconstructions are illustrated on Fig.~\ref{fig:datasets}.

\section{Related work}
\label{sec:related}

Surface reconstruction has been extensively studied from 3D points \cite{bergerhal01017700} and/or images \cite{FurukawaCGV2015}.  We consider here
the input to be 3D line segments (with viewpoints), that can be sparse, missing, noisy and corrupted with outliers.  We aim at an idealized piecewise-planar and watertight surface.

\textbf{To deal with sparse data}, some methods detect planes based on 3D features and dominant orientations \cite{SinhaICCV2009}, possibly with a Manhattan-world assumption \cite{FurukawaCVPR2009}, and create piecewise-planar depth maps taking into account visibility and local consistency. %
Other approaches consider 2D image patches back-projected on 3D planes \cite{MicusíkIJCV2010,BourkiWACV2017}.  In contrast, our method produces a watertight mesh, does not impose a few specific orientations, and can work with 3D features only, not requiring  images and not working at pixel level.\\
\indent
Another approach to little input data is to extend boundaries and primitives until they intersect \cite{castellani2002improving}. It however does not ensure a watertight reconstruction either. This is only achieved by methods that create a volumetric partition of the 3D space and extract the surface from full and empty cells. The partition can be a voxel grid \cite{schnabel2009completion}, a 3D Delaunay triangulation \cite{LabatutICCV2007,wan2011variational} or a plane arrangement \cite{Chauve:cvpr2010,Boulch:sgp2014}. %

\textbf{Wireframe reconstruction} is what most lined-based methods focus on: rather than surfaces, they study how to generate meaningful 3D line segments \cite{Jain:cvpr2010, HoferBMVC2013, Hofer:cviu2016, IenagaACCVw2017}, after line matching is performed \cite{SchmidCVPR1997}.  And more general curves than lines are not used beyond structure from motion \cite{NurutdinovaICCV2015}.

\textbf{Surface reconstruction with lines in addition to points} has received a modest attention.
\cite{BaillardCVPR1999} reconstruct planes from a 3D line and a neighboring detected point. It requires lines surrounded with texture and is outlier-sensitive. It also does not prevent self-intersections nor guarantees watertightness.
\cite{Bay3DPVT2006} segments images into likely planar polygons based on 3D corner junctions and use best supporting lines to reconstruct polygons in 3D.
For 2.5D reconstruction, extracted 3D lines \cite{SchmidCVPR1997} are used with a dense height map to build a line arrangement on the ground plane and create geometric primitives and building masks~\cite{ZebedinECCV2008}.
In \cite{SinhaICCV2009}, pairs of 3D lines generated from vanishing directions provide plane hypotheses, validated by 3D points. The surface is a set of planar patches created from plane assignment to pixels.
\cite{Sugiura3DV2015} adds points uniformly sampled on the 3D lines to the Delaunay triangulation, introducing extra parameters, and although visibility is treated without sampling, the method is unlikely to work on scenes with only sparse lines. \cite{Hofer3DV2014} also shows a meshing improvement using 3D line segments.

\textbf{Surface reconstruction from line segments only}, when points fail due to the lack of texture, has little been studied.\\
\cite{ZaheerECCV2012} presents a single-view surface reconstruction based on 2D line segments. 
Lines are paired from segment extensions along their direction, and planes orientations are sought by RANSAC, hypothesizing mutually orthogonal corresponding 3D lines. Articulating lines are found at plane intersections to construct a multi-plane structure.
Our \emph{structural lines} are called `articulation' or `articulating' lines in \cite{ZaheerECCV2012}. They are discovered late, to set plane offsets, whereas we differentiate them early at plane detection.
For robotic mapping, \cite{WittICRA2014} considers all combinations of two non-collinear coplanar line segments as plane hypotheses. Line segments are then assigned to possibly multiple planes in a face complex built from plane intersections. The reconstructed surface is made of faces depending on an occlusion score.
Compared to our approach, this method does not scale well to many lines, is sensitive to outliers, relies on a number of conservative heuristics that can be detrimental to surface recall, involves no regularization, and does not reconstruct a watertight mesh.
As for \cite{MentgesICRA2016}, it first reprojects 3D lines into images that see them, studies the intersection of segments in 2D rather than planes in 3D, and infers plane hypotheses. The surface is made from image faces back-projected onto a possible 3D plane.  Although less sensitive to outliers, this method involves heuristics and no proper regularization, and it reconstructs a non-watertight mesh with floating polygons and possible self-intersections.

\textbf{Extracting 3D planes from line segments} has little been treated; the literature focuses on point clouds, chiefly ignoring line clouds.  The most popular scheme for points, which is robust to sparsity contrary to region growing as in \cite{Chauve:cvpr2010,Boulch:sgp2014}, is RANSAC \cite{ChoiBMVC2009, schnabel-2007-efficient}.
But as explained below, it cannot straightforwardly be applied to line segments because it relies on different distribution hypotheses and because of the possible association of a segment to several primitives, also invalidating line discretization into points. Still, \cite{WittICRA2014} takes line segments as input, but plane detection is somewhat exhaustive, hence with scalability issues, and sensitive to outliers.  Using laser data, \cite{CABO201528} exploits 3D lines to detect planes, but it uses strong properties of lidar acquisition, namely line parallelism and large and dense data.\\
\indent
An open question is if multi-model methods \cite{ZulianiICIP2005,ToldoECCV2008,Isack2012,MagriCVPR2014}, which assume non-overlapping segmented data, can be adapted not only to large inputs but also to multiple shape support \cite{KriegelTKDD2009,BaadelSAI2016}, as absolutely required for line segments.

\textbf{Surface reconstruction from a plane arrangement} is a common topic, with variants enforcing plane regularity \cite{LiSIGGRAPH2011,MonszpartEtAl:RAPter:2015} or level of detail \cite{verdie:hal-01113078}, or offering reconstruction simplicity \cite{NanICCV2017}. It is largely orthogonal to our work. Here we build on~\cite{Boulch:sgp2014}, with line-specific data and visibility terms.

\section{Plane detection from 3D line segments}
\label{sec:detection}

The first step of our approach is to detect planes that are supported by line segments in the input line cloud~$\Linesegs$. 
We use the RANSAC framework \cite{Fischler:1981:RSC:358669.358692} as it scales well to large scenes and deals well with a high proportion of outliers, which are unavoidable in photogrammetric data.

As argued above and shown experimentally (cf.\ Sect.~\ref{sec:expe}), a key requirement is to allow a line to belong to two planes.  Lines supporting one plane are considered \emph{textural}; lines supporting two planes are deemed \emph{structural}. Yet some actual texture lines may support additional ``virtual'' planes, as when a line is drawn around an object, e.g., at the borders of a frieze around the walls of a room, which belongs both to the vertical walls and to an non-physical horizontal plane.

\paragraph*{Candidate plane construction.}
We generate candidates by sampling the minimum number of observations required to create a model, i.e., two non-collinear line segments to define a plane.
Two 3D segments $\lineseg_a, \lineseg_b$ can be coplanar in two ways: they can be parallel, or their supporting infinite lines can intersect. 
With noisy real data, the latter can be relaxed using a maximum small distance $\epsilon$ between the lines.
We discard parallelism because, when reconstructing man-made environments such as buildings, it may generate many bad planes. Indeed, two random vertical segments (e.g., detected on windows) are parallel but statistically unlikely to support an actual, physical plane (e.g., segments on different facades). We thus threshold the angle $\angle (\lineseg_a,\lineseg_b)$, which also excludes the degenerate case of collinear segments.

\paragraph*{Greedy detection and multi-support issues.}

We sample planes as line pairs and perform an iterative extraction of the most significant planes, i.e., with the largest number of supporting segments after a given number of sampling trials.
However, contrary to usual RANSAC, we cannot remove supporting  segments at once as they may actually belong to two planes; it would lead to detecting the main planes only, missing planes with a smaller support.
The supplementary material (supp.\,mat.) illustrates failure cases.
Conversely, we cannot consider all segments as available at each iteration: it would statistically lead to multiple detections of the same large planes and again miss planes with small support.

A natural way to allow a datum to be part of several detection in greedy RANSAC is to remove inliers for model sampling but not for data assignment to models \cite{ZaheerECCV2012}. But for sparse data (which is the case with line segments), it fails to detect models with little data support, e.g., preventing detecting all the faces of a cube from its sole edges.

Another way to allow the same datum to seed several models is to bound their number, i.e., 2 for lines supporting planes. But it does not work either as it often associates a line twice to more or less the same plane. As illustrated in the supplementary material too, this yields very bad results. %

Our solution is to bound the number of supported planes per line segment, but with an additional condition to prevent shared segments to belong to similar planes.

\paragraph*{Candidate plane generation.}

We note $\Linesegsof(\plane)$ the set of line segments supporting a plane~$\plane$, $\Planesof(\lineseg)$ the set of planes supported by a line segment $\lineseg \in \Linesegs$, with $\card{\Planesof(\lineseg)} \leq 2$, and $\Linesegs_i$ the set of segments supporting $i$~plane(s) for $i$ in $0,1,2$. 

We construct these sets iteratively by generating candidates planes $\plane$ and assigning them segments $\lineseg \in \Linesegs$, some of which may have already been assigned to another plane $\Planesof(\lineseg)$. Only line segments in $\Linesegs_2$ are discarded from the pool of available segments to support a plane, as they already support two planes. Initially, $\Linesegs_0 = \Linesegs$, and $\Linesegs_1 = \Linesegs_2 = \emptyset$.

As line segments are not put aside as soon as they are assigned to a plane, they can be drawn again to generate new candidate models.  However, generating several times the same plane (with the same supporting line segments) would not only reduce efficiency, but also make some models little likely to be drawn, as models with a large support would be sampled much more often.  To prevent it, after drawing a first line segment $\lineseg_a \in \Linesegs_0 \cup \Linesegs_1$, there are two cases.
If $\lineseg_a \in \Linesegs_0$, i.e., if $\lineseg_a$ has not been assigned to any plane yet, then the second segment $\lineseg_b$ can be drawn unconditionally in $\Linesegs_0 \cup \Linesegs_1$ as it will always yield a new model.
If $\lineseg_a \in \Linesegs_1$, i.e., if $\lineseg_a$ has already been assigned to some plane $\altplane$, with $\Planesof(\lineseg_a) = \{\altplane\}$, then lines in $\Linesegsof(\altplane)$, i.e., supporting~$\altplane$, are excluded when drawing the second segment $\lineseg_b$. This ensures $\lineseg_a,\lineseg_b$ cannot participate to the same already existing model.
As the number of extracted planes is typically less than a few hundred, this drawing can be optimized by incrementally keeping track of the sets $\bar\Linesegsof(\plane) = \Linesegs \setminus (\Linesegs_2 \cup \Linesegsof(\plane))$, that have \emph{not}  already been assigned to a detected plane $\plane$.

We do not prevent a line pair to be redrawn when it previously failed to generate an accepted model (for lack of planarity, parallelism or poor support) because it does not lead to unbalanced chances to detect a plane.  And if $\card{\Linesegs}$ is not too large, we can draw systematically all line pairs.

Note that we do not prevent a line pair to be redrawn when it previously failed to generate an accepted model (for lack of planarity, parallelism or poor support).
It is not an issue as it does not lead to unbalanced chances to detect a plane. Yet, when the number of input line segments is not too large, we can perform a systematic drawing of all line pairs, possibly exploiting the above filtering.  In this case, all possible models are considered and at most once.

\paragraph*{Inlier selection.}

After picking a candidate plane $\plane$, we populate the support $\Linesegsof(\plane)$.  For this, we go through each segment $\lineseg \in \Linesegs_0 \cup \Linesegs_1$ and assign it to $\Linesegsof(\plane)$ if close enough to $\plane$, i.e., if $d(\lineseg,\plane) \leq \epsilon$. Several distances can be used, such as the average or the maximum distance to the plane.

If $\lineseg$ already supports some other plane $\plane'$, i.e., if $\Planesof(\lineseg) = \{\plane'\}$, then also assigning $\lineseg$ to $\plane$ would make it a structural segment. As such, we impose that it lies close to the line at the intersection of both planes, i.e., $d(\lineseg,\plane \cap \plane') \leq \epsilon$.
This condition is stronger than imposing both $d(\lineseg,\plane) \leq \epsilon$ and $d(\lineseg,\plane') \leq \epsilon$ as the angle between $\plane$ and $\plane'$ could be small and $\lineseg$ could then be close to both $\plane$ and $\plane'$ although far from their intersection. This condition is actually crucial. Without it, we would tend to associate $l$ to two planes $P$ and $P'$ which are very similar, and fail to detect crease lines.

\paragraph*{Plane selection.} Last, we sample  $N_\iiter$ models and keep the plane with the largest number of inliers. (See the supp.\,mat.\ for the abstract version of the algorithm.)

This plane detection differs from \cite{ZaheerECCV2012}, that samples and populates planes from 2D line pairs instead of 3D lines, making inlier search quadratic, not linear, and requiring heuristically to only consider pairs defined by intersecting segment extensions, which is highly unstable due to noise in endpoints and which induces plane splitting at occlusions.
We have none of these downsides.
Besides, structural lines in \cite{ZaheerECCV2012} are found with heuristics after RANSAC, considering plane pairs and candidate lines, which only makes sense as they have few ($<$10) planes.  We get them directly, without heuristics, in greater number, and for many more planes.

\paragraph*{Plane refitting.}

After each plane $\plane_\ibest$ is selected, it is actually refitted to its inliers $\Linesegsof_\ibest$ before being stored into~$\Planes$, based on the (signed) distance of the segment endpoints, weighted by the segment length. As it changes the plane equation, we check whether the slice centered on the refitted plane $\plane'$ with thickness $\epsilon$ now contains extra segments.  If so, they are added as inliers and refitting is repeated.

\paragraph*{Plane fusion.}

Modeling a building may require different levels of details, including small plane differences such as wall offsets for door jambs, baseboards or switches. But setting a small $\epsilon$ to do so may easily break a wall or a ceiling into several fragments because it is not perfectly planar due to construction inaccuracies or load deflections. Each country actually has standards (official or not) defining construction tolerances, e.g., 1\,cm error every 2\,m for walls.

To prevent this arbitrary fragmentation while preserving details, we add a plane fusion step with a tolerance higher than $\epsilon$, i.e., with a maximal distance threshold $\epsilon_\ifus > \epsilon$ to the plane refitted on the union of inliers.  This allows merging at $\epsilon_\ifus$ accuracy several plane fragments detected at $\epsilon$. However, to make sure it applies only to cases described above, we impose a maximum angle $\theta_\ifus$ when merging two planes and minimum proportion $p_\ifus$ of common inliers.  Concretely, we consider all pairs of planes in $\Planes$ whose angle is less than $\theta_\ifus$, sort them, pick the pair with the smallest angle, and try merging it.  If it succeeds, the two planes are removed, the new refitted plane is added, and the priority queue based on angles is updated before iterating.
If it fails, the pair of planes is discarded and the next pair is considered.
This is similar to a heuristics used in Polyfit \cite{NanICCV2017}.

\paragraph*{Plane limitation.} To make sure not too many planes are given to the surface reconstruction step, because of possible limitations (cf.\ Sect.~\ref{sec:conclu}), the algorithm may be stopped after at most $N_\imax$ (best) greedy detections.

\section{Surface reconstruction}
\label{sec:surface}

The second step of our approach is surface reconstruction based on detected planes and observations of 3D line segments.
Rather than selecting plane-based faces with hard constraints for the the surface to be manifold and watertight \cite{NanICCV2017}, we follow \cite{Chauve:cvpr2010,Boulch:sgp2014} and consider a scene bounding box, partition it into 3D cells constructed from the planes, and assign each cell with a status `full' or `empty' depending on segment visibility, with a regularization prior coping with sparse and missing data. The reconstructed surface is then the interface between full and empty cells. By construction, it is watertight and free from self-intersections.
Our contribution is a total reformulation of \cite{Chauve:cvpr2010,Boulch:sgp2014} in terms of lines, making the difference between textural and structural lines, and with a lighter treatment of noise in data.

The volume partition is given by a cell complex $\Cells$ made from an arrangement of planes detected in the line cloud.
For each cell $\cell \in \Cells$, we represent occupancy by a discrete variable $\occup_\cell \in \{0,1\}$: $0$ for empty and $1$ for full.
A surface is uniquely defined by a cell assignment $\voccup : \Cells \mapsto \{0,1\}$, where $\voccup(\cell) = \occup_\cell$.  
The optimal cell assignment $\voccup$ is defined as the minimum of an energy $E(\voccup)$ which is the sum of three terms: a primitive term $E_\iprim(\voccup)$ penalizing line segments not lying on the reconstructed surface, a visibility term $E_\ivis(\voccup)$ penalizing surface reconstructions on the path between observations and their viewpoints, and a regularization term $E_\iregul(\voccup)$ penalizing complex surfaces. 
\begin{equation}
E(\voccup) = E_\iprim(\voccup) + E_\ivis(\voccup) + E_\iregul(\voccup)
\end{equation}

\paragraph*{Dealing with noise.}

To deal with noise in input data, \cite{Boulch:sgp2014} introduces slack in the choice of cells penalized for not being at the reconstructed surface and lets regularization make the right choices. The resulting formulation and resolution is heavy.
Instead, we assume that plane extraction (Sect.~\ref{sec:detection}) did a good-enough job: any segment supporting a plane (resp.\ two planes) is considered as a noisy inlier and is projected on the plane (resp.\ the intersection of the two planes). A segment not supporting any plane is treated as an outlier for data fidelity (no penalty for not being on the reconstructed surface) but not for visibility (penalty for not being seen from viewpoints if hidden by reconstructed surface).

\paragraph*{Primitive term.}

$E_\iprim(\voccup)$ penalizes line segments that support planes but do not lie on the reconstructed surface. But it does not penalize the presence of matter in front of segments w.r.t.\ viewpoints, letting the visibility term do it.
Segments that support no plane are ignored as if outliers.

For a segment $\lineseg$ supporting one plane $\plane$, and for each viewpoint $\viewpoint$ seeing at least a part of $\lineseg$, we consider the set $C$ of all cells $\cell$ immediately behind $\lineseg$ w.r.t.\ $\viewpoint$, possibly only along a fraction $\lineseg_\cell$ of $\lineseg$ due to occlusions (cf.\ Fig.~\ref{fig:energy}(a)).
Each $\cell\,{\in}\, C$ is penalized if not full, with a cost $1{-}\occup_\cell$.

For a segment $\lineseg$ supporting two planes $\plane_1,\plane_2$, a cell behind $\lineseg$ w.r.t.\ viewpoint $\viewpoint$ need not be full. (Penalizing emptiness actually yields terrible results, as the supp.\,mat.\ shows.) 
Any configuration is valid as long as the space around $\lineseg$ is not empty (cf.\ Fig.~\ref{fig:energy}(b)): salient edges, reentrant edges or planes (if the seemingly structural line happens to only be textural).
To penalize only when all three cells $\cell$ around a visible fraction of $\lineseg$ are empty (ignoring the cell in front), we consider a cost of $\max(0,1-\sum_\cell\occup_\cell)$, which is equal to~$1$ in this case, and $0$ in other configurations. %

Both textural and structural cases can be covered with a single formula, where we weigh the cost by the length of the visible fraction of $\lineseg$ and normalize it by a scale of interest~$\sigma$\rlap:
\vspace*{-1.5mm}
\begin{equation}
E_\iprim(\voccup)\!=\!
\sum_{\mathclap{\lineseg \in \Linesegs_1 \cup \Linesegs_2}} ~~~~
\sum_{\viewpoint \in \Viewpoints(\lineseg)\vphantom{\lineseg}} ~~~~
\sum_{\mathclap{C \in \Cells(\lineseg, \viewpoint)}} ~~
\frac{\len{\lineseg_C}}{\sigma}\!
\max(0, 1\!-\!\sum_{\mathclap{\substack{\cell \in C}}} \occup_\cell)
\label{eq:eprim}
\end{equation}
where $\Linesegs_1{\cup}\Linesegs_2$ is the set of segments $\lineseg$ supporting at least one plane, $\Viewpoints(\lineseg)$ is the set of viewpoints $\viewpoint$ seeing $\lineseg$, $\Cells(\lineseg, \viewpoint)$ is the set of cells $\cell$ adjacent to $\lineseg$ but not in the triangles of sight from $\viewpoint$ to non occluded fragments of $\lineseg$ (locally 1 or 3 cells as to whether $\lineseg$ belongs to 1 plane or 2 planes), $\lineseg_C$ is the set of fragments of $\lineseg$ in each cell $c \in C$, and $\len{\lineseg_C}$ is the sum of the lengths of segment fragments in $\lineseg_C$. %

\begin{figure}[t]
\vspace*{-2mm}
\centering
\begin{minipage}[b]{0.46\linewidth}
\centering
\includegraphics[width=\linewidth]{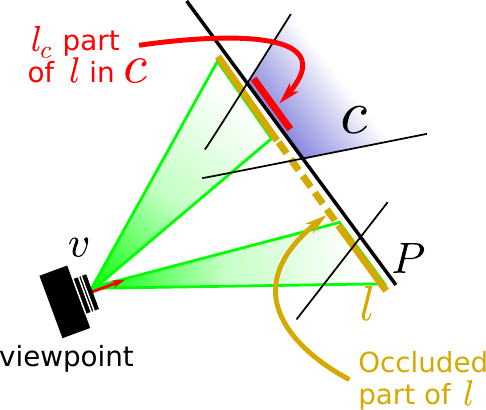}\\ \vspace{0.1cm}
(a) Primitive term, $\lineseg \in \plane$
\end{minipage}
\hfill
\begin{minipage}[b]{0.52\linewidth} 
\centering
\includegraphics[width=\linewidth]{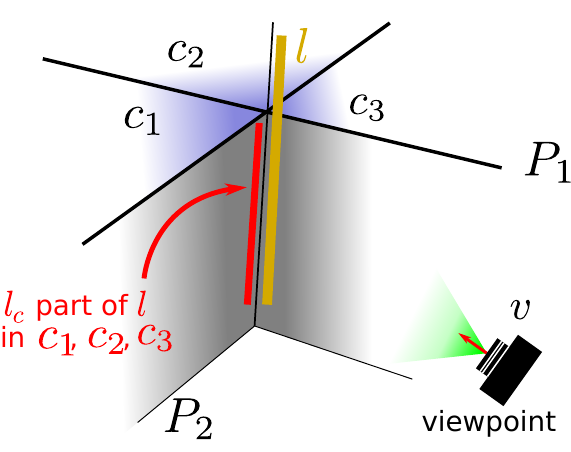}\\ \vspace{0.1cm}
(b) Primitive term, $\lineseg \in \plane_1 \cap \plane_2$
\end{minipage}

\vspace*{3mm}
\begin{minipage}[b]{\linewidth}
\centering
\includegraphics[width=0.9\linewidth]{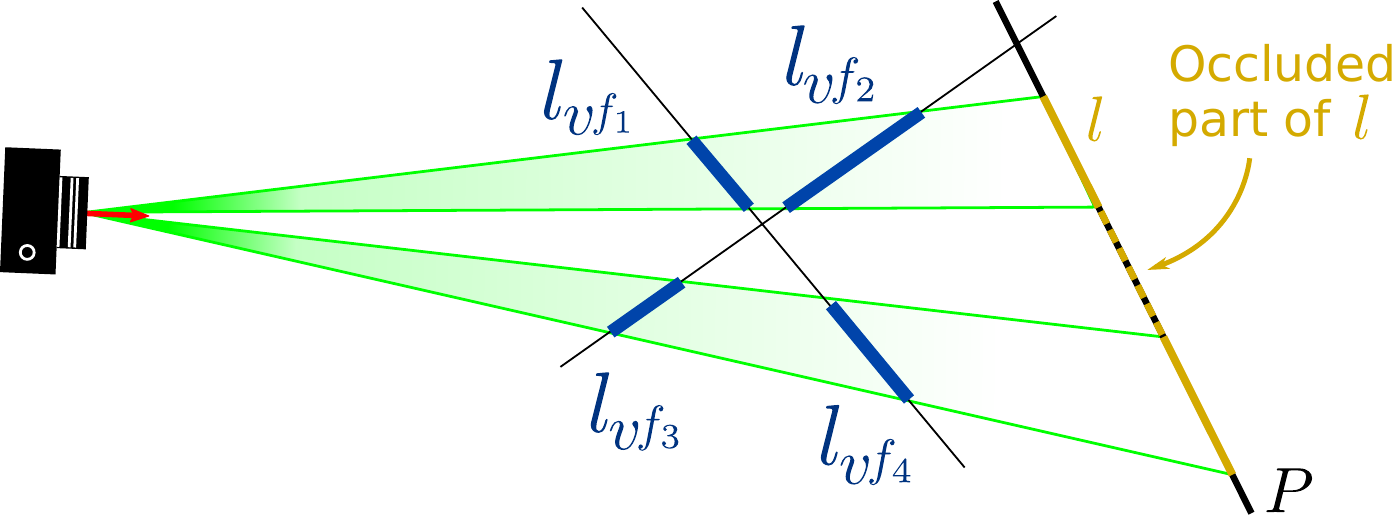} %
\end{minipage}
\vspace*{-6mm}
\caption{~~(c) Visibility term}%
\label{fig:energy}
\vspace*{-4mm}
\end{figure}

\begin{figure*}
\noindent
\adjustbox{valign=b}{\begin{tabular}{@{}c@{~}c@{~}c}
\includegraphics[width=0.22\linewidth]{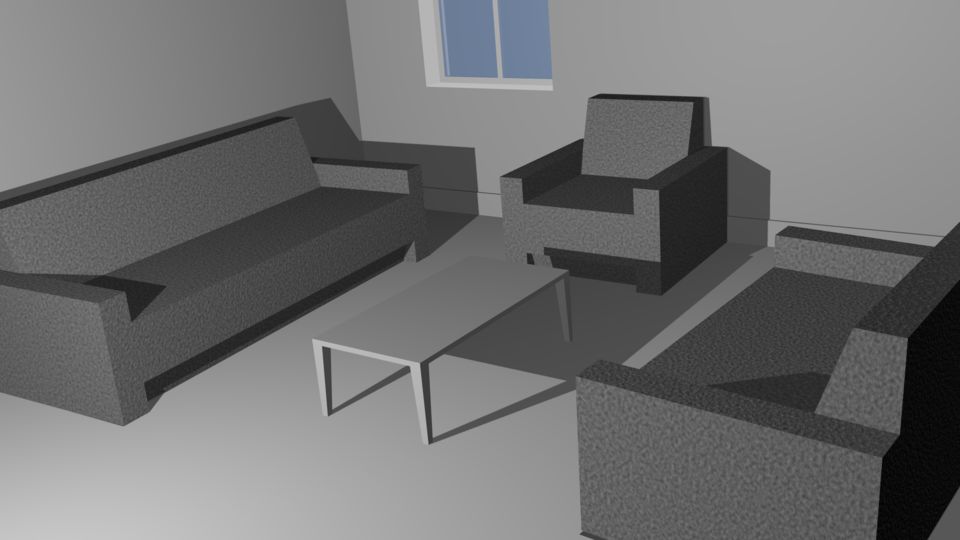}\raisebox{1mm}{\llap{(1)\hspace*{33mm}}} &
\includegraphics[width=0.205\linewidth, trim={0 2cm 0 0}, clip]{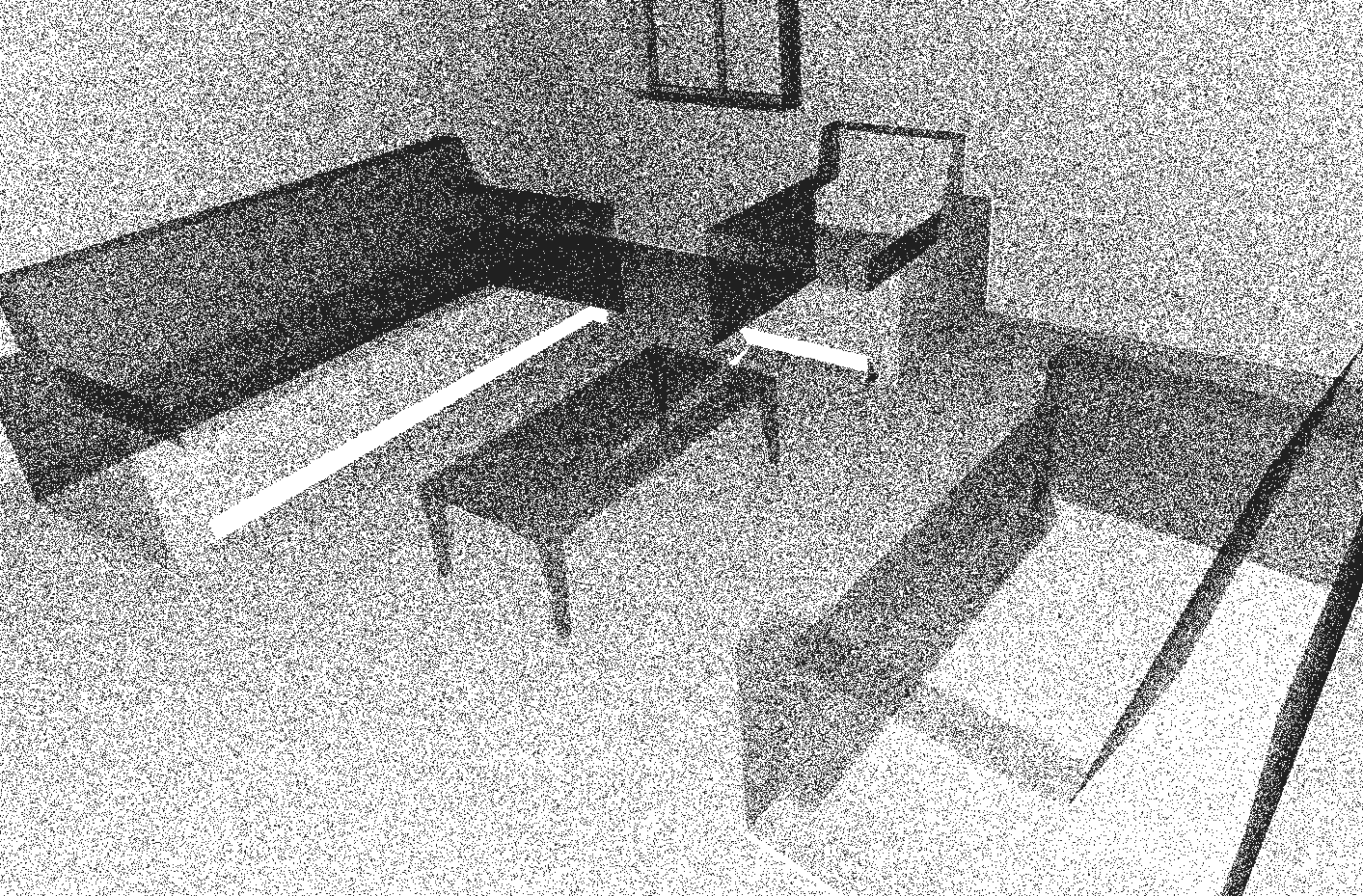}\raisebox{1mm}{\llap{(2)\hspace*{31mm}}} &
\includegraphics[width=0.205\linewidth, trim={0 3cm 0 0}, clip]{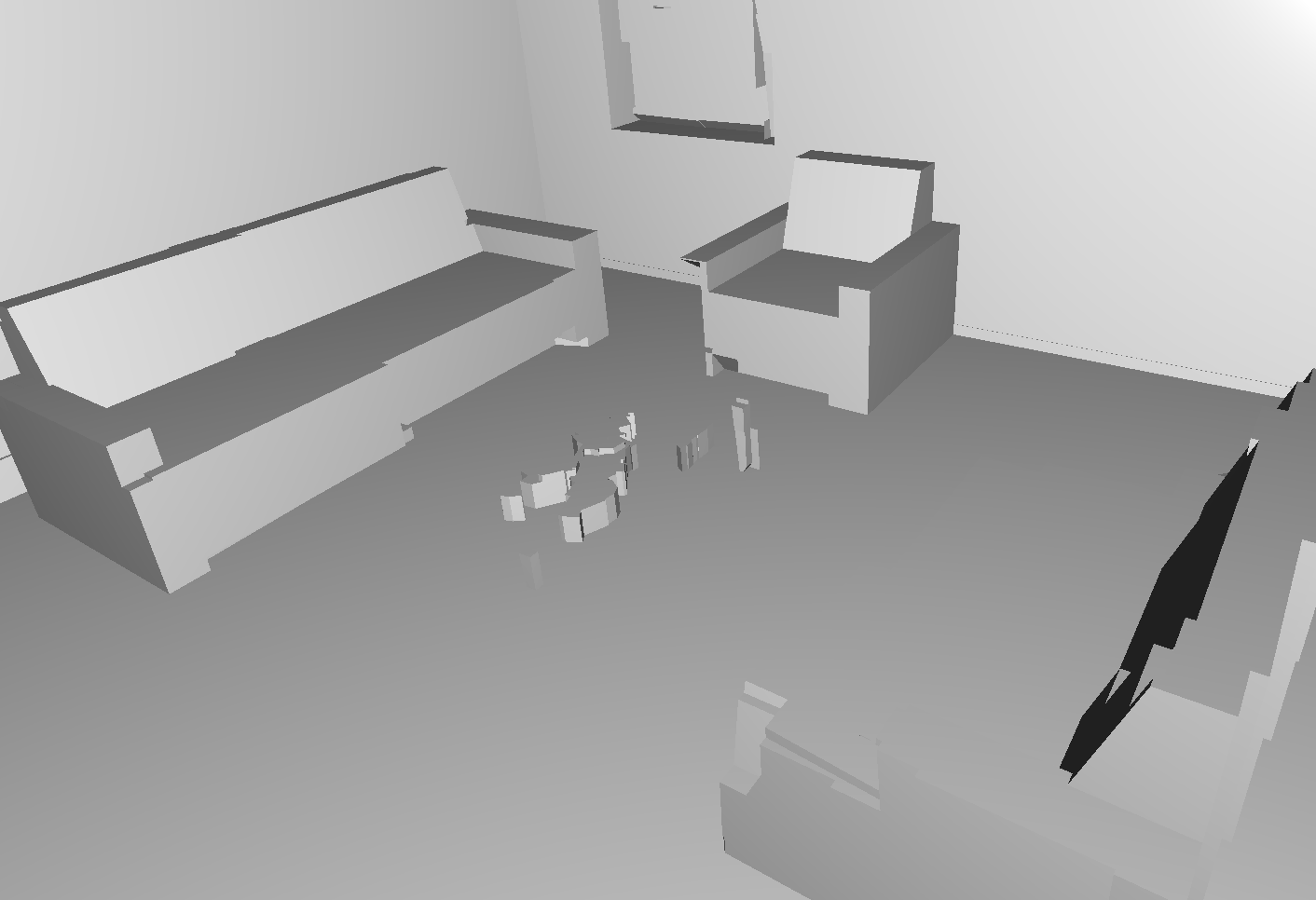}\raisebox{1mm}{\llap{(3)\hspace*{31mm}}}
\\
\includegraphics[width=0.22\linewidth, trim={0 6cm 0 0}, clip]{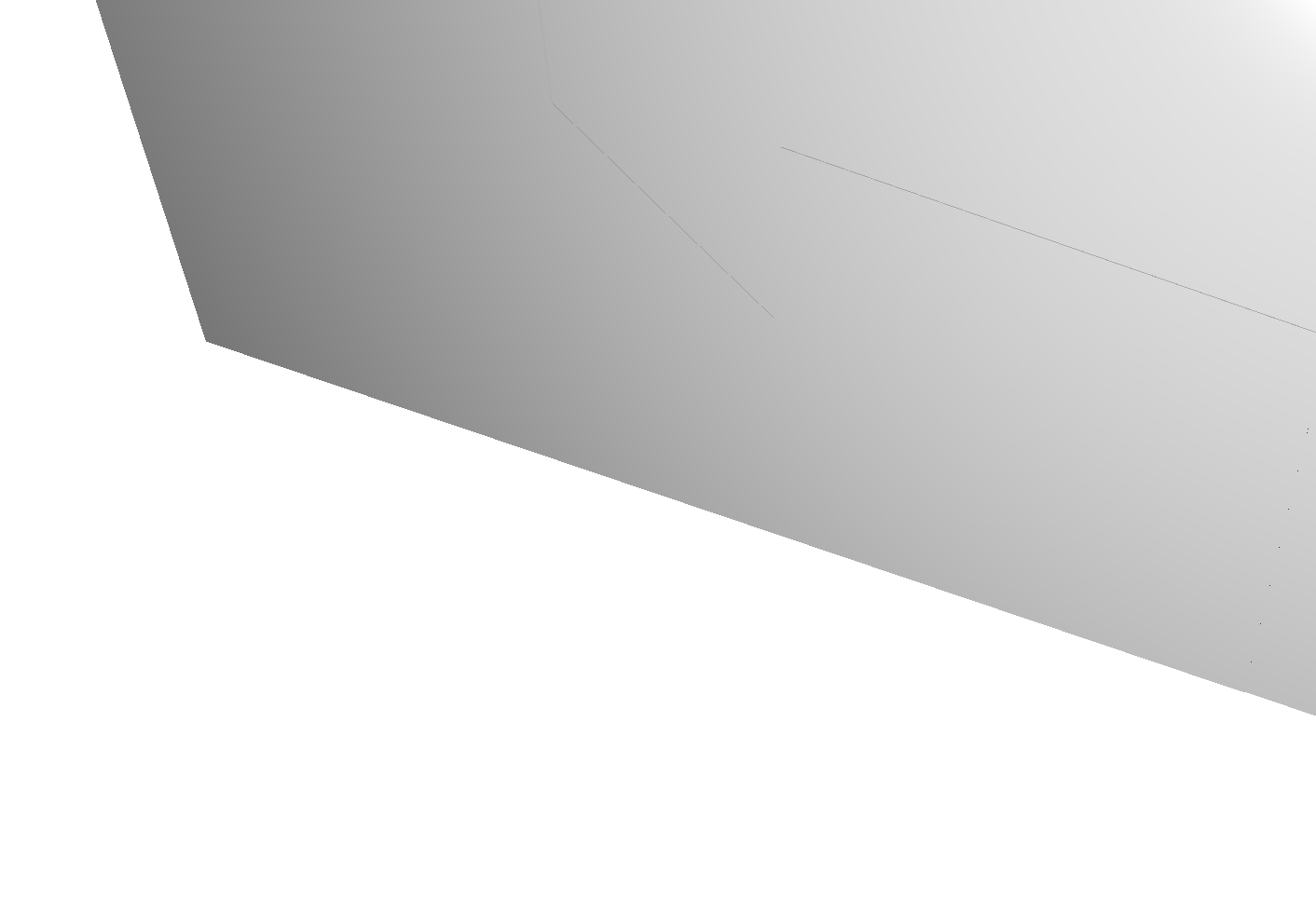}\raisebox{1mm}{\llap{(4)\hspace*{33mm}}} &
\includegraphics[width=0.205\linewidth]{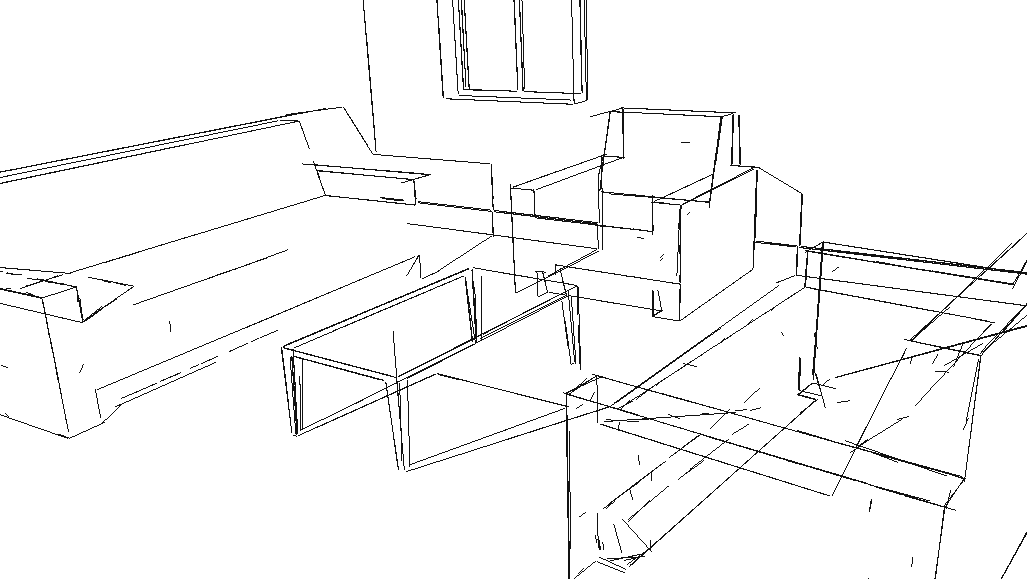}\raisebox{1mm}{\llap{(5)\hspace*{31mm}}} &
\includegraphics[width=0.205\linewidth]{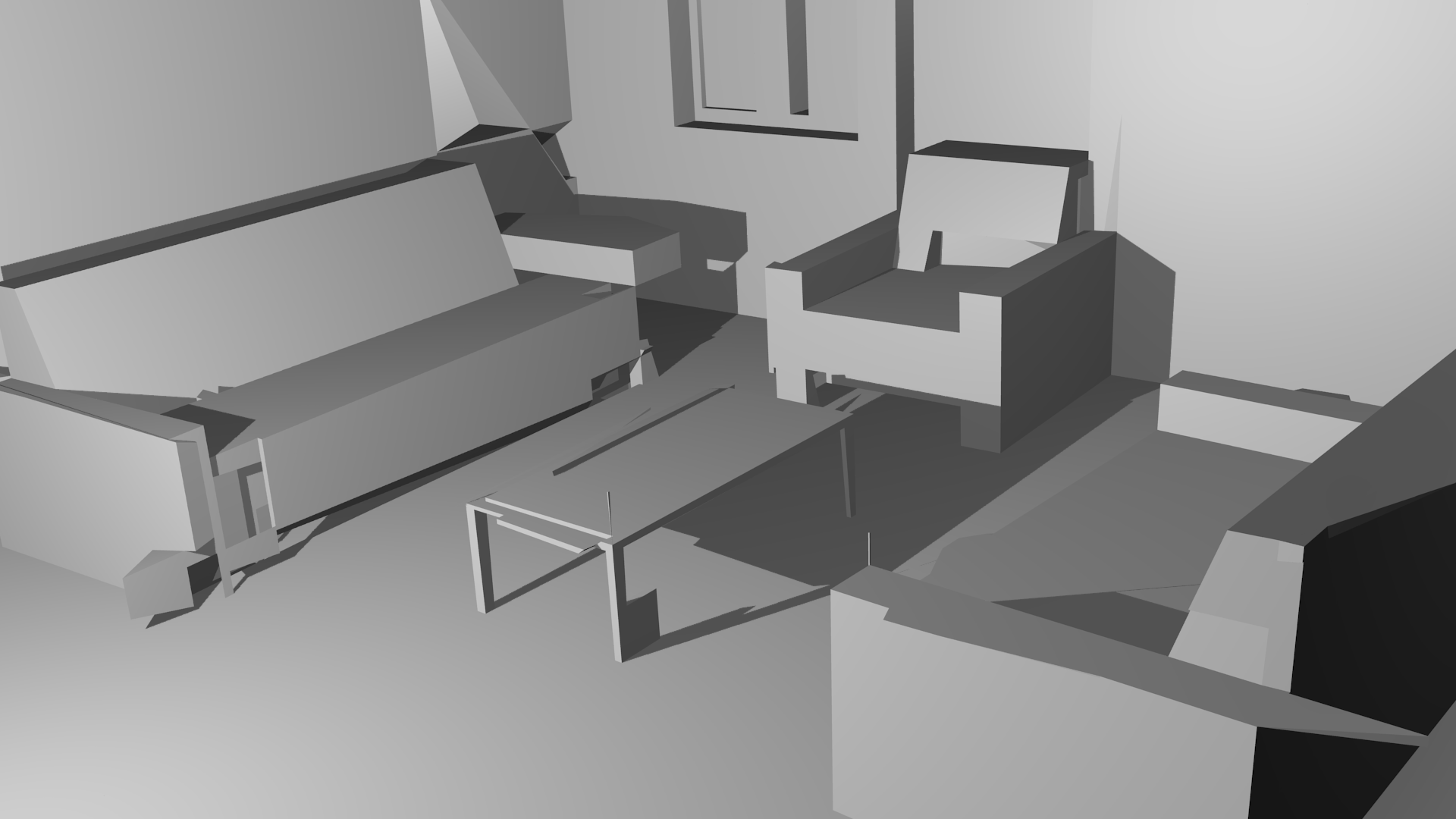}\raisebox{1mm}{\llap{(6)\hspace*{31mm}}}
\end{tabular}}\raisebox{1.8mm}{\includegraphics[width=0.342\textwidth, trim={0 3mm 0 0},clip]{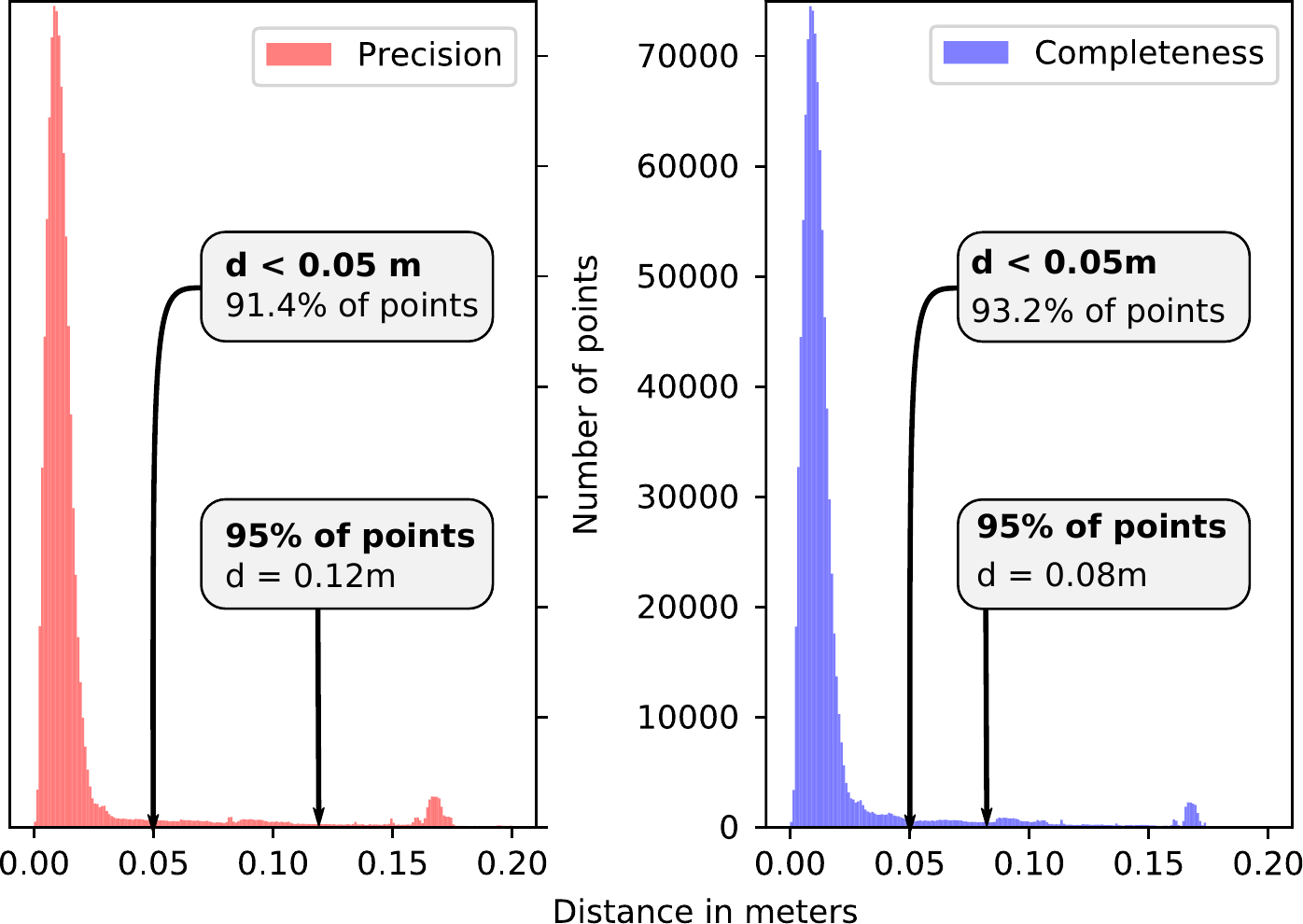}}\raisebox{0mm}{\llap{(7)\hspace*{28mm}}}
\vspace{-3mm}
\caption{\inhouse{}: (1)~an image of the dataset, (2)~points densely sampled on surface, (3)~reconstruction with~\cite{Chauve:cvpr2010}, (4)~failed reconstruction with \cite{Chauve:cvpr2010} from points sampled on lines, (5)~3D lines detected with Line3D++~\cite{Line3Dpp}, with noise and outliers, (6)~our reconstruction, which is nonetheless superior, (7)~histograms of distance errors w.r.t. ground truth (m).}
\label{fig:baseline}
\vspace*{-4mm}
\end{figure*}

\paragraph*{Visibility term.}

$E_\ivis(\voccup)$ penalizes reconstructed surface boundaries between viewpoints and segments, as \cite{Chauve:cvpr2010,Boulch:sgp2014}. It measures the number of times a 3D segment $\lineseg$ is to be considered an outlier as it should not be visible from a given viewpoint $\viewpoint$, weighted by the length of the visible parts $\lineseg_{\viewpoint,\face}$ of $\lineseg$ on the offending faces $\face$ (possibly fragmented due to occlusions). Contrary to $E_\iprim(\voccup)$, all segments are considered in $E_\ivis(\voccup)$, not just segments supporting a plane: 
\vspace*{-1.5mm}
\begin{equation}
E_\ivis(\voccup)\!=\!\lambda_\ivis \sum_{\lineseg \in \Linesegs} \, \sum_{\viewpoint \in \Viewpoints(\lineseg)} \,
\sum_{\substack{\face \in \Faces(\lineseg, \viewpoint) }} \frac{\len{\lineseg_{\viewpoint,\face}}}{\sigma} \, \abs{\occup_{\cell_\face^{{+}\viewpoint}} - \occup_{\cell_\face^{{-}\viewpoint}}}
\end{equation}
where $\Faces(\lineseg, \viewpoint)$ is the set of faces $\face$ of the complex intersected by the visibility triangle $(\lineseg, \viewpoint)$, at some unoccluded segment fractions $\lineseg_{\viewpoint,\face}$ totalizing a length of $\len{\lineseg_{\viewpoint,\face}}$, and $\cell_\face^{{+}\viewpoint},\cell_\face^{{-}\viewpoint}$ are the cells on each side of $\face$ ($\smash{\cell_\face^{{+}\viewpoint}}$ being nearest to $\viewpoint$).

\paragraph*{Regularization term.}

$E_\iregul(\voccup)$ penalizes surface complexity as the sum of the length of reconstructed edges and the number of corners, with relative weights $\lambda_\iedge, \lambda_\icorner$, as defined in \cite{Boulch:sgp2014}. Area penalization makes little sense here due to the low density of observations in some regions.

\paragraph*{Solving.}
Minimizing this higher-order energy is a non linear integral optimization problem ($\max$ in eq.~\eqref{eq:eprim}). 
As in \cite{Boulch:sgp2014}, integral variables are relaxed to real values and slack variables are introduced.
The resulting linear problem is solved and fractional results are rounded to produce the final integral values.
See details in the supplementary material.
\paragraph*{Properties of reconstructed surface.}

By construction, the surface we produce is watertight, even if the input data is very sparse, and not self-intersecting.  Our process treats outliers (with RANSAC at plane detection stage, and regularization during reconstruction) and noise (with a model tolerance at plane detection stage and via projections when reconstructing).  It has also several positive properties:\\
\mbox{}\quad $\bullet$\hspace{1.5mm}\textit{Insensitivity to line over-segmentation:} if a 3D line segment $\lineseg$ is split, $E(\voccup)$ does not change and thus the same surface is reconstructed. This provides robustness to over-segmentation, which is a common weakness of line segment detectors. (It may however change inlier-ness.)\\
\mbox{}\quad $\bullet$\hspace{1.5mm}\textit{Little sensitivity at endpoints}: given a line segment $\lineseg$, slightly changing its endpoint %
only makes a marginal change to $E(\voccup)$.  (Yet it may change inlier-ness too.)\\
\mbox{}\quad $\bullet$\hspace{1.5mm}\textit{Insensitivity to dummy planes:} given a 3D cell assignment~$\voccup$, if an extra plane is randomly inserted in the arrangement, the value of $E_\ivis(\voccup)$ does not change as it only depends on surface transitions encountered on visibility path.

\section{Experiments}
\label{sec:expe}

We experimented both with real and synthetic data,
for qualitative and quantitative analysis.
The real datasets consist of images of a `MeetingRoom' from \cite{Salaun3DV2017}, of a `Barn' from Tanks and Temples~\cite{Knapitsch2017}, of a `DeliveryArea', a 'Bridge' and of a corridor named `Terrains' from ETH3D~\cite{schoeps2017cvpr}. All scenes are poorly textured (walls of uniform colors). %
The synthetic datasets include a `TimberFrame' house \cite{Jain:cvpr2010} as well as two new synthetic datasets, to be publicly released. `\inhouse{}' is a living room, with both large planar areas (walls, floor and ceiling) and smaller details (chair and table legs). `Andalusian' is the outside of a modern house; it is piecewise-planar and uniformly white. %

\begin{table}[t]
    \centering\setlength{\tabcolsep}{3pt}
    \vspace*{-01mm}
    \begin{tabular}{|c|c|c|c|c|c|c|c|c|c|}
         \hline
         ${\sigma_p}^\dagger$ & $\epsilon$ & $\epsilon_\ifus$ & $\theta_\ifus$ & $p_\ifus$ & $N_\iiter$ & $N_\imax$ & $\lambda_\ivis$ & $\lambda_\iedge$ & $\lambda_\icorner$ \\
         \hline
         \small 2.5 &
         \small 2\,cm &
         \small 3\,$\epsilon$ &
         \small 10\textdegree &
         \small 20\% &
         \small 50k &
         \small 160 &
         \small 0.1 &
         \small 0.01 &
         \small 0.01 \\
         \hline
    \end{tabular}
    \vspace*{-2.5mm}
    \caption{Parameters (all datasets are metric). $\dagger$\,in Line3D++}
    \label{tab:def_params}
\vspace{1.5mm}
\small
\setlength\tabcolsep{2pt}
\begin{tabular}{|l|cccccccc|}
\hline
Dataset & \#img & $\card{\Linesegs}$ & $\card{\Planes}$ & $\card{\Planes_{\ifus}}$ & $\card{\Linesegs_0}$ & $\card{\Linesegs_1}$ & $\card{\Linesegs_2}$ & \#res \\
\hline
TimberFrame & 241 & 7268 & 140 & 131 & 264 & 4507 & 2497 & 79024 \\
Andalusian & 249 & 1234 & 160 & 148 & 242 & 597 & 395 & 14503 \\
MeetingRoom & 32 & 831 & 135 & 130 & 25 & 383 & 423 & 9028 \\
Terrains & 42 & 3223 & 120 & 105 & 9 & 356 & 2858 & 18189 \\
DeliveryArea & 948 & 1586 & 160 & 160 & 30 & 771 & 785 & 29222 \\
Barn & 410 & 7936 & 160 & 141 & 41 & 2157 & 5738 & 83989 \\
Bridge & 110 & 7437 & 150 & 102 & 338 & 4168 & 2931 & 48315 \\
\inhouse{} & 159 & 1995 & 120 & 106 & 1 & 286 & 1708 & 18304 \\
\hline
\end{tabular}
\vspace*{-2.5mm}
\caption{Dataset statistics: number of images \#img, number of 3D line segments~$\card{\Linesegs}$, number of 3D planes before fusion~$\card{\Planesof}$, number of 3D planes after fusion~$\card{\Planesof_{\ifus}}$, number of segments supporting no plane~$\card{\Linesegs_0}$, one plane~$\card{\Linesegs_1}$ or two planes~$\card{\Linesegs_2}$, and total number of sub-segments \#res.}
\label{tab:stat} 
\vspace*{-3.5mm}
\end{table}

MeetingRoom was calibrated with LineSfM~\cite{LineSfM,Salaun3DV2017} and we recalibrated the other real datasets using Colmap \cite{SchoenbergerCVPR2016}, with distortion correction as it impacts line detection.  The synthetic datasets came with their exact calibration. 

We then ran Line3D++ \cite{Line3Dpp}, as defined in \cite{Hofer:cviu2016}, to detect and reconstruct 3D line segments. As seen on Figs.~\ref{fig:datasets}, \ref{fig:baseline}, \ref{fig:colmap} and in the supplementary material, line segments obtained from Line3D++ are extremely noisy: lines that are mostly parallel, orthogonal, planar or colinear in the original scene turn out to be reconstructed with visible discrepancies.  There are also many missing lines and many outliers.  For instance, in MeetingRoom, many segments are floating in the air in the middle of the room.  Line3D++ also tends to duplicate the same segment many times with a little displacement, leading to a treatment as noise or outlier.

\begin{figure*}[!t]
\centering
\begin{tabular}{c@{~}c@{~}c@{~}c}
\multicolumn{4}{c}{
    \begin{tabular}{c@{~}c@{~}c}
        \includegraphics[width=0.255\linewidth]{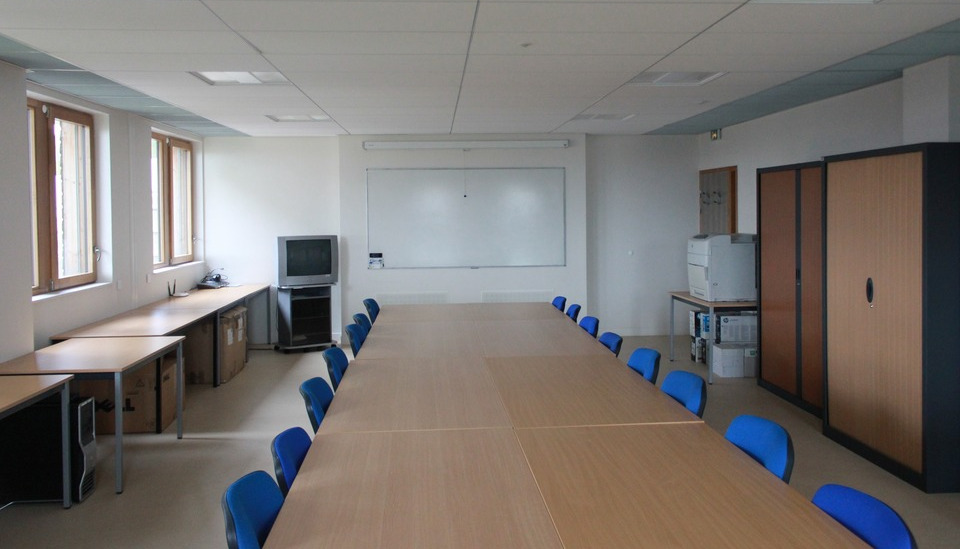}\raisebox{1mm}{\llap{(a1)\hspace*{19mm}}} &
        \includegraphics[width=0.255\linewidth]{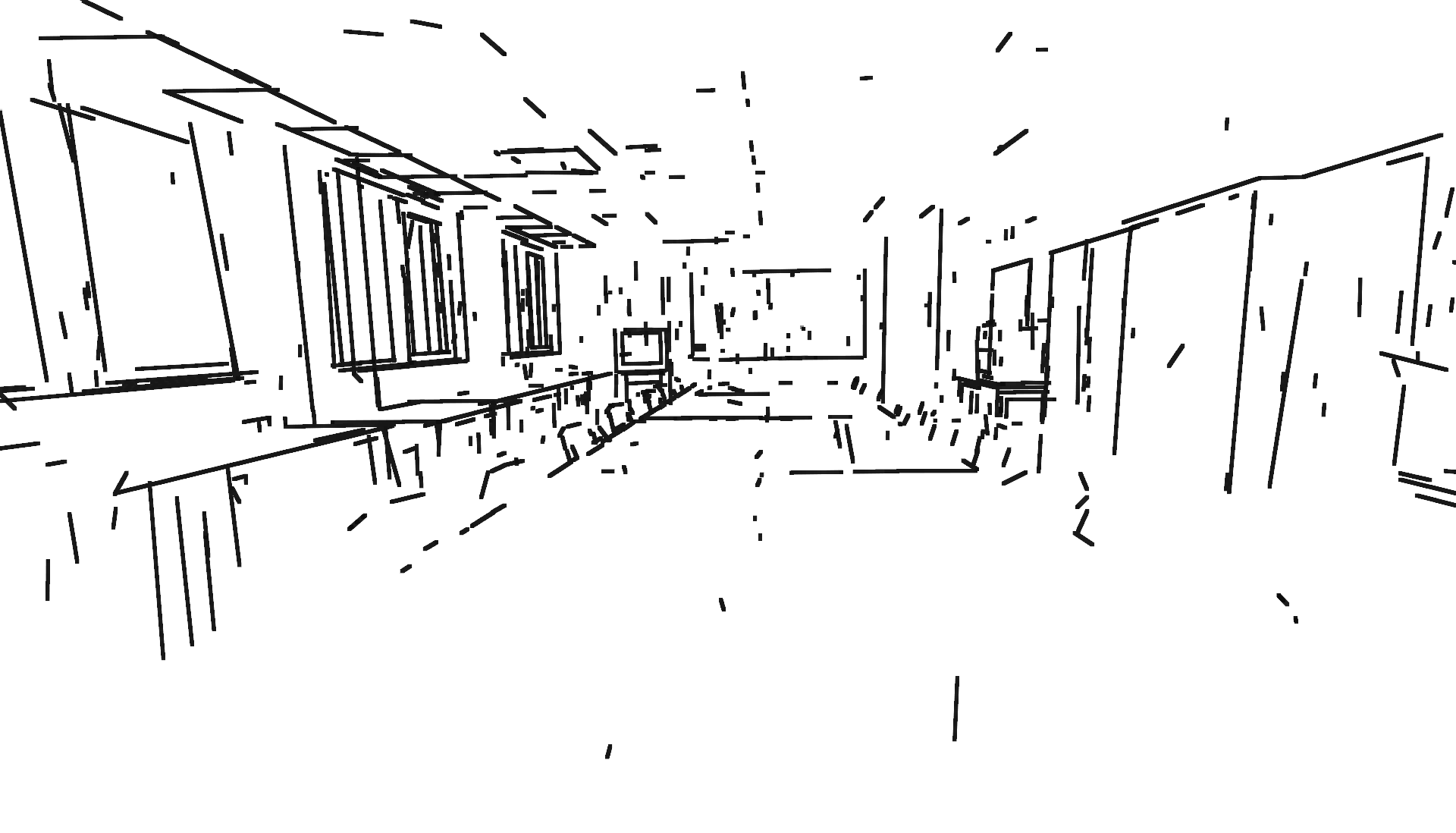}\raisebox{1mm}{\llap{(a2)\hspace*{19mm}}} &
        \includegraphics[width=0.255\linewidth]{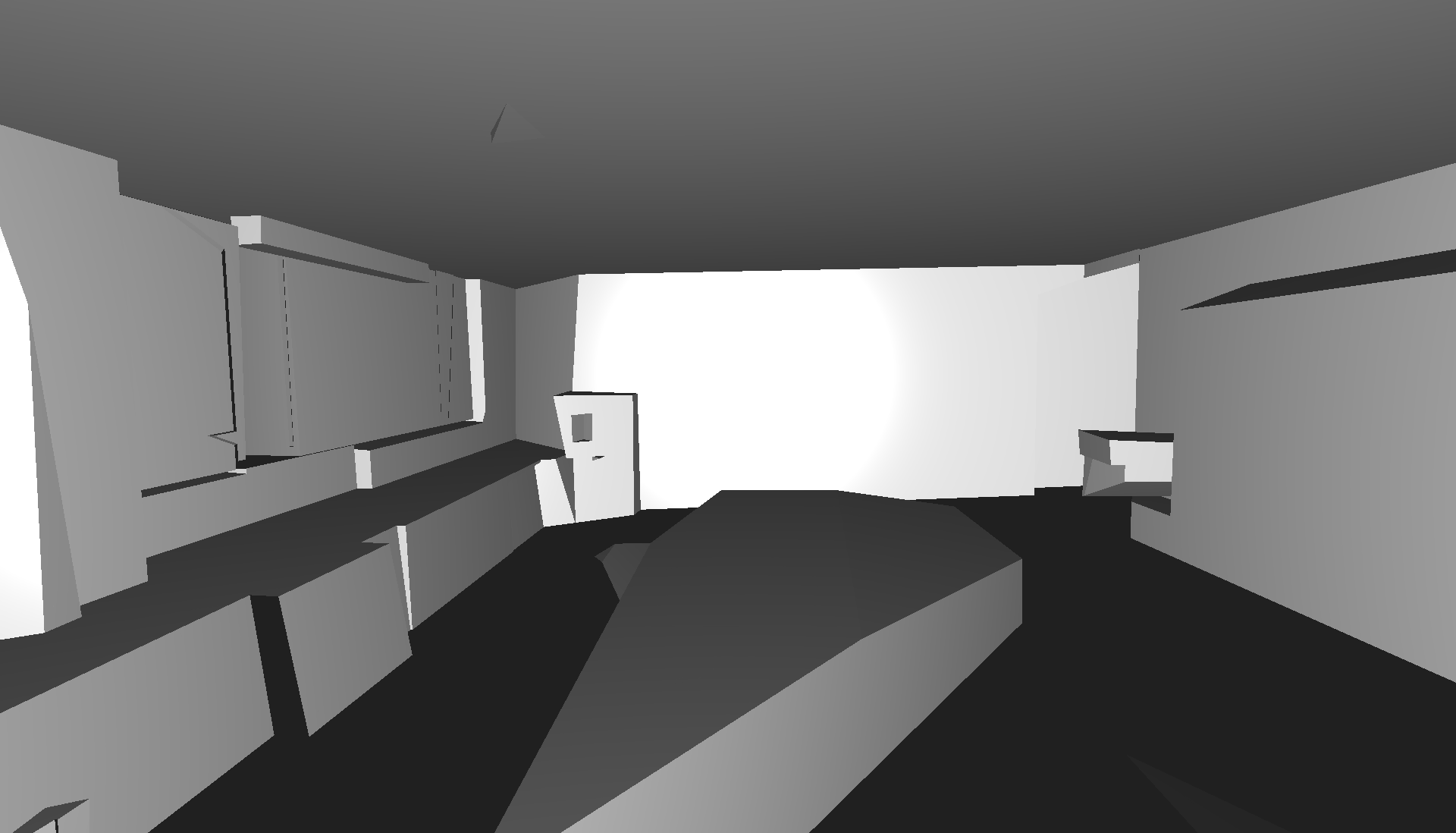}\raisebox{1mm}{\llap{(a3)\hspace*{19mm}}}
    \end{tabular}
} \\
\includegraphics[width=0.24\linewidth]{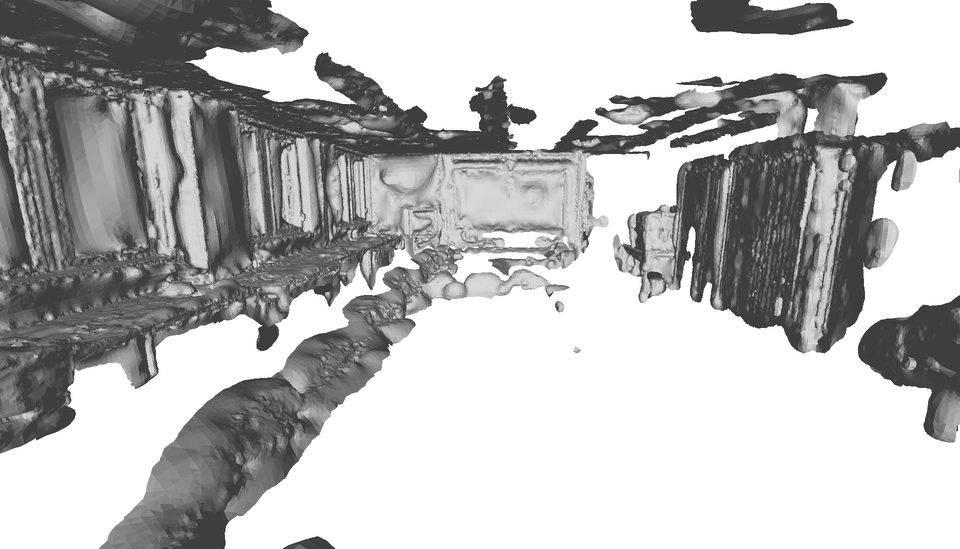}\raisebox{1mm}{\llap{(a4)\hspace*{20mm}}} &
\includegraphics[width=0.24\linewidth]{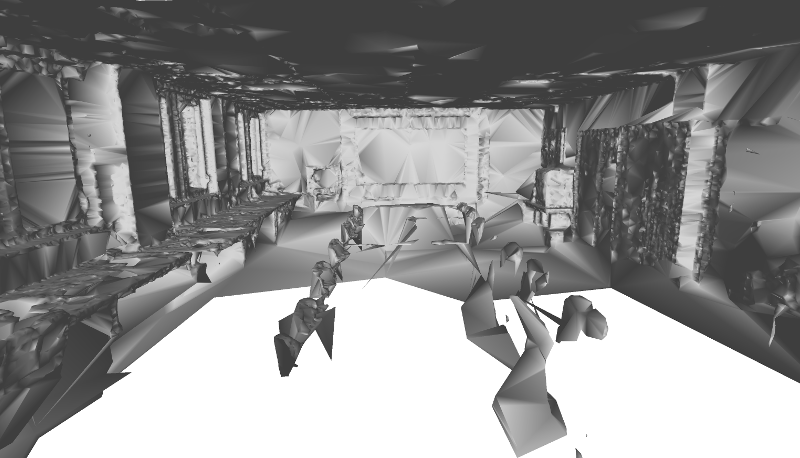}\raisebox{1mm}{\llap{(a5)\hspace*{20mm}}} &
\includegraphics[width=0.24\linewidth]{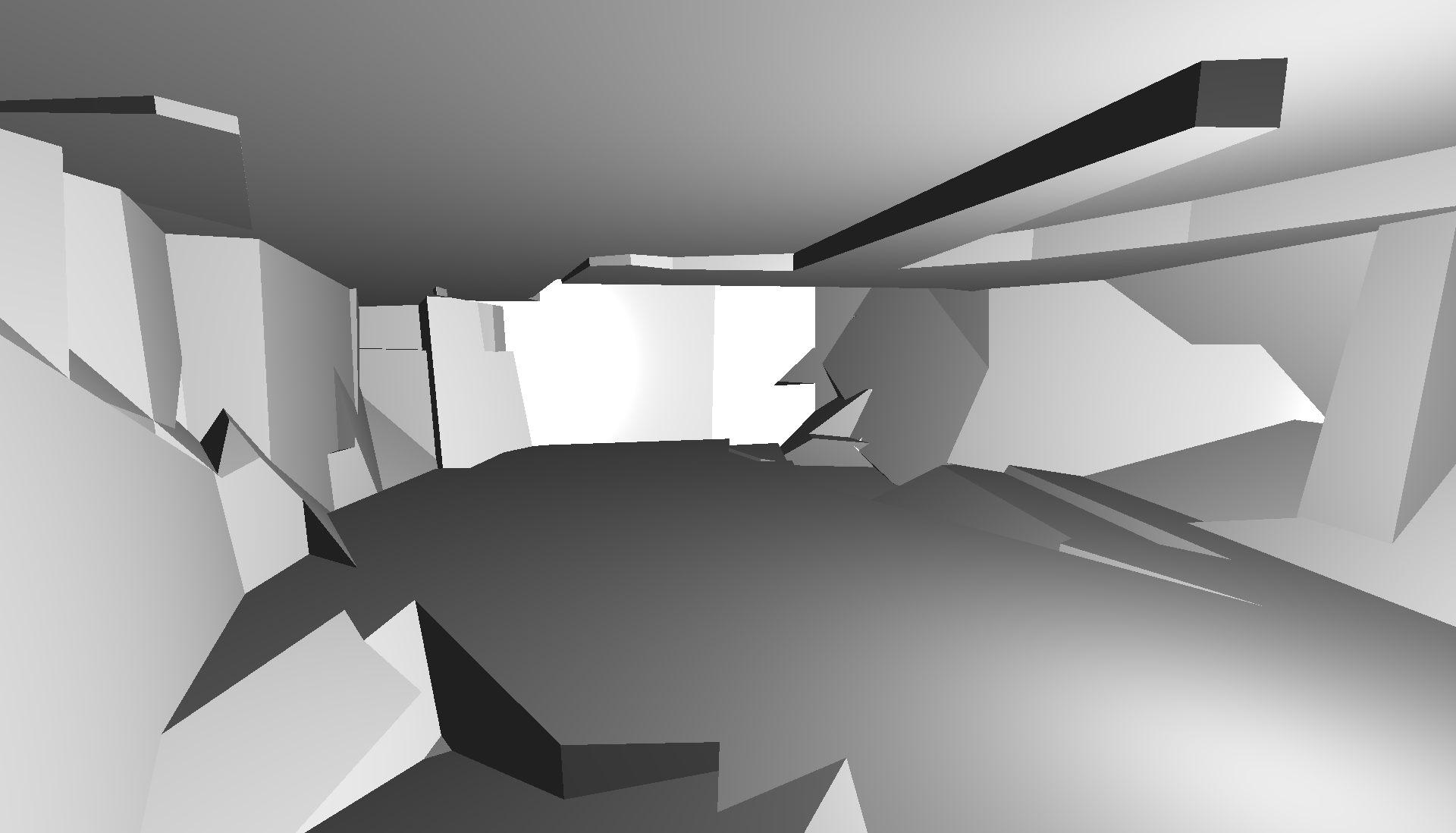}\raisebox{1mm}{\llap{(a6)\hspace*{10mm}}} &
\includegraphics[width=0.24\linewidth]{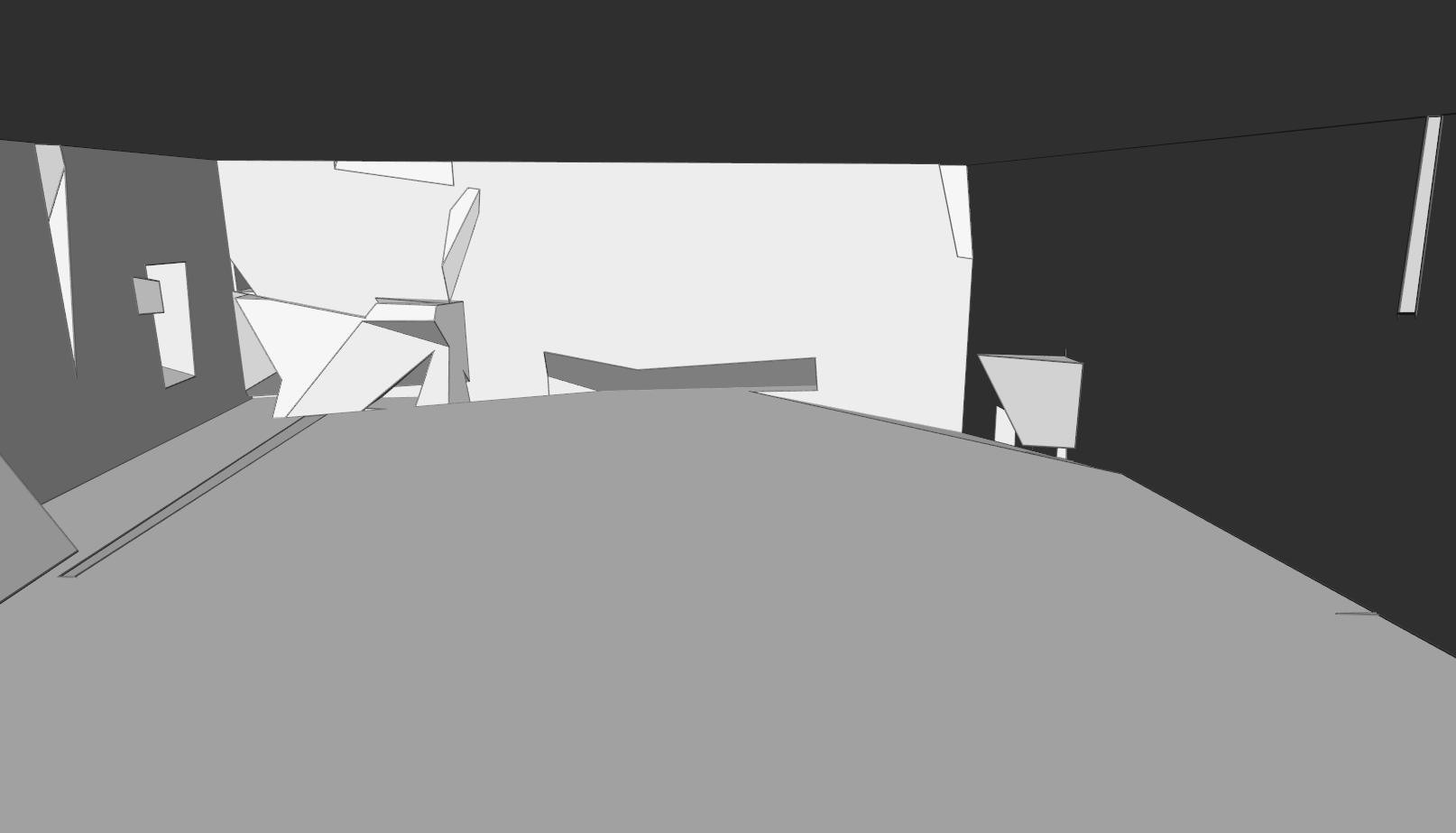}\raisebox{1mm}{\llap{(a7)\hspace*{10mm}}} \\
\hline \\[-0.9em]
\multicolumn{4}{c}{
    \begin{tabular}{c@{~}c@{~}c}
    \includegraphics[width=0.255\linewidth]{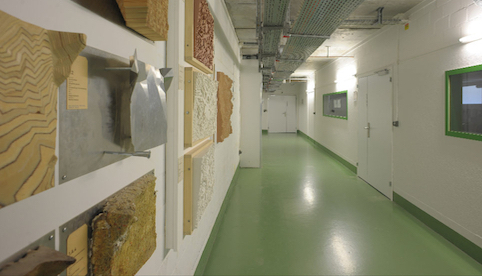}\raisebox{1mm}{\llap{(b1)\hspace*{19mm}}} &
    \includegraphics[width=0.255\linewidth]{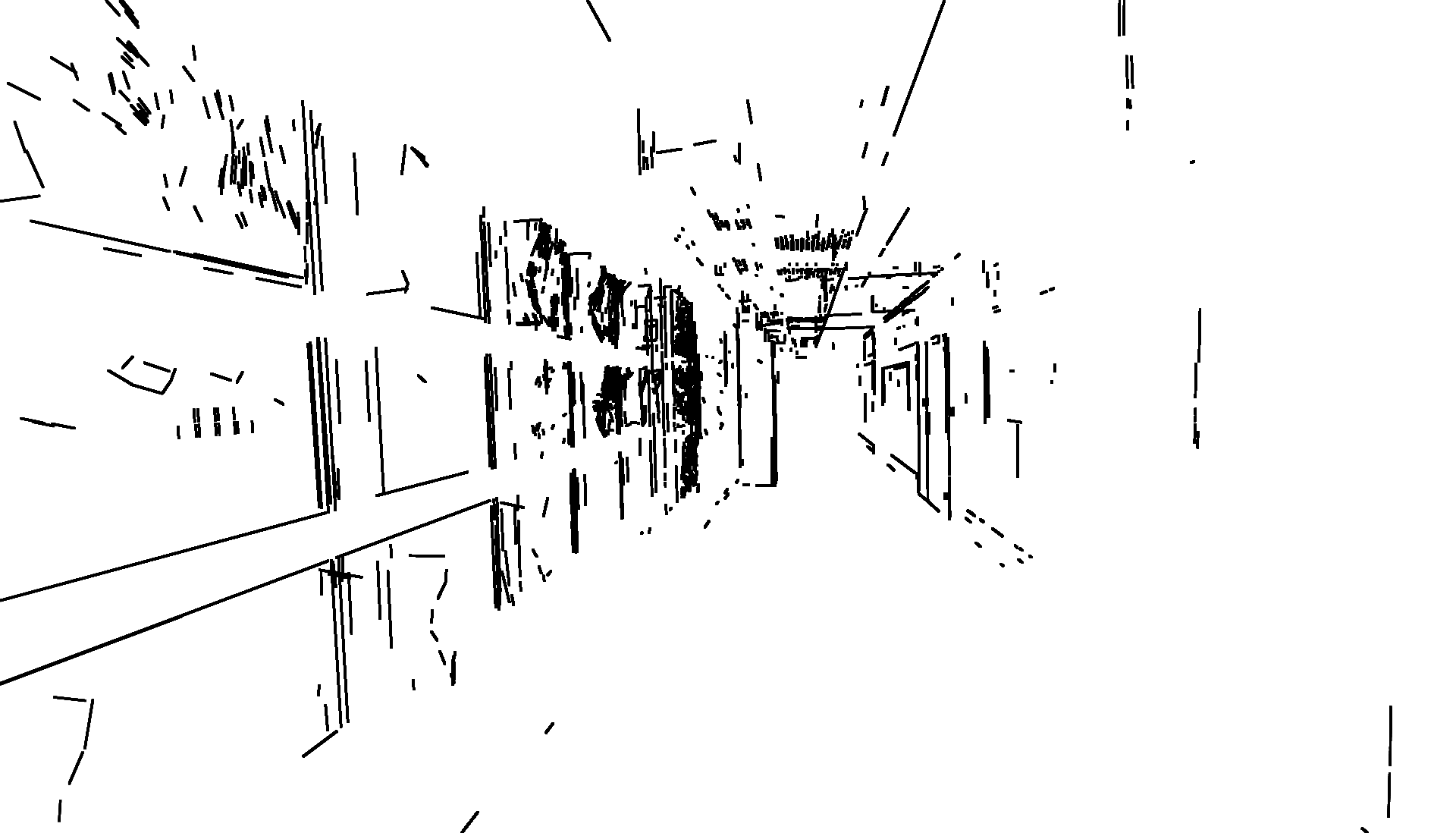}\raisebox{1mm}{\llap{(b2)\hspace*{19mm}}} &
    \includegraphics[width=0.255\linewidth]{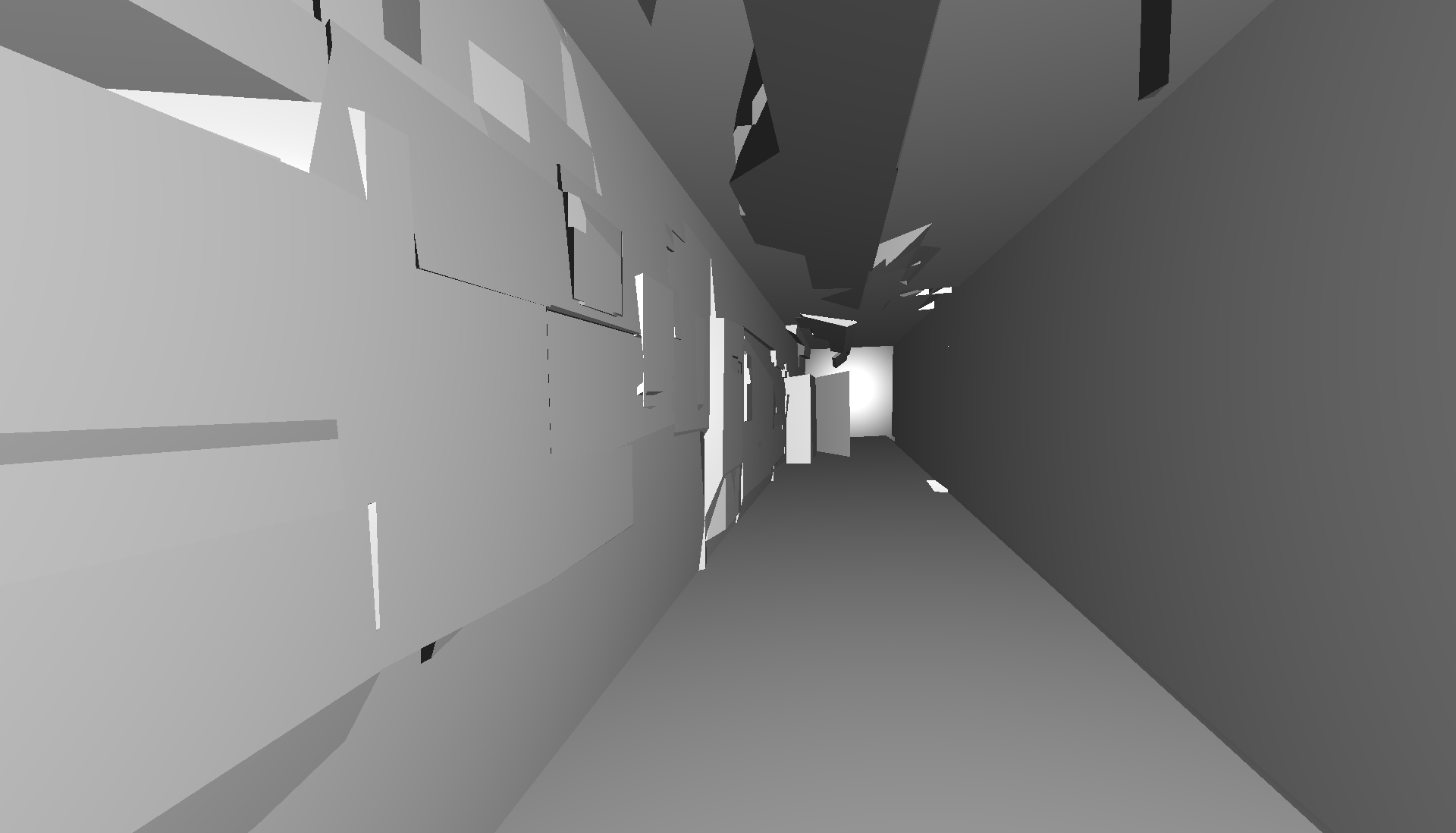}\raisebox{1mm}{\llap{(b3)\hspace*{19mm}}}
    \end{tabular}
} \\
\includegraphics[width=0.24\linewidth]{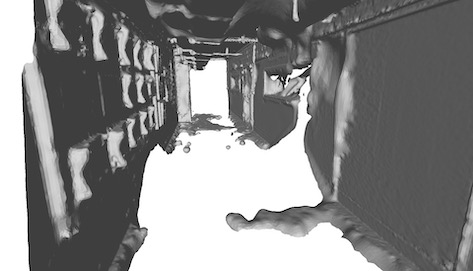}\raisebox{1mm}{\llap{(b4)\hspace*{20mm}}} &
\includegraphics[width=0.24\linewidth]{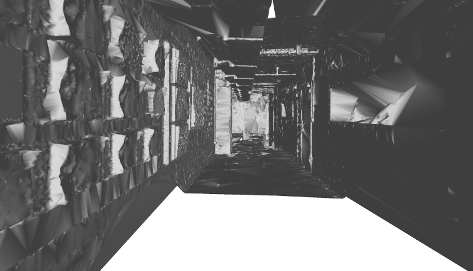}\raisebox{1mm}{\llap{(b5)\hspace*{20mm}}} &
\includegraphics[width=0.24\linewidth]{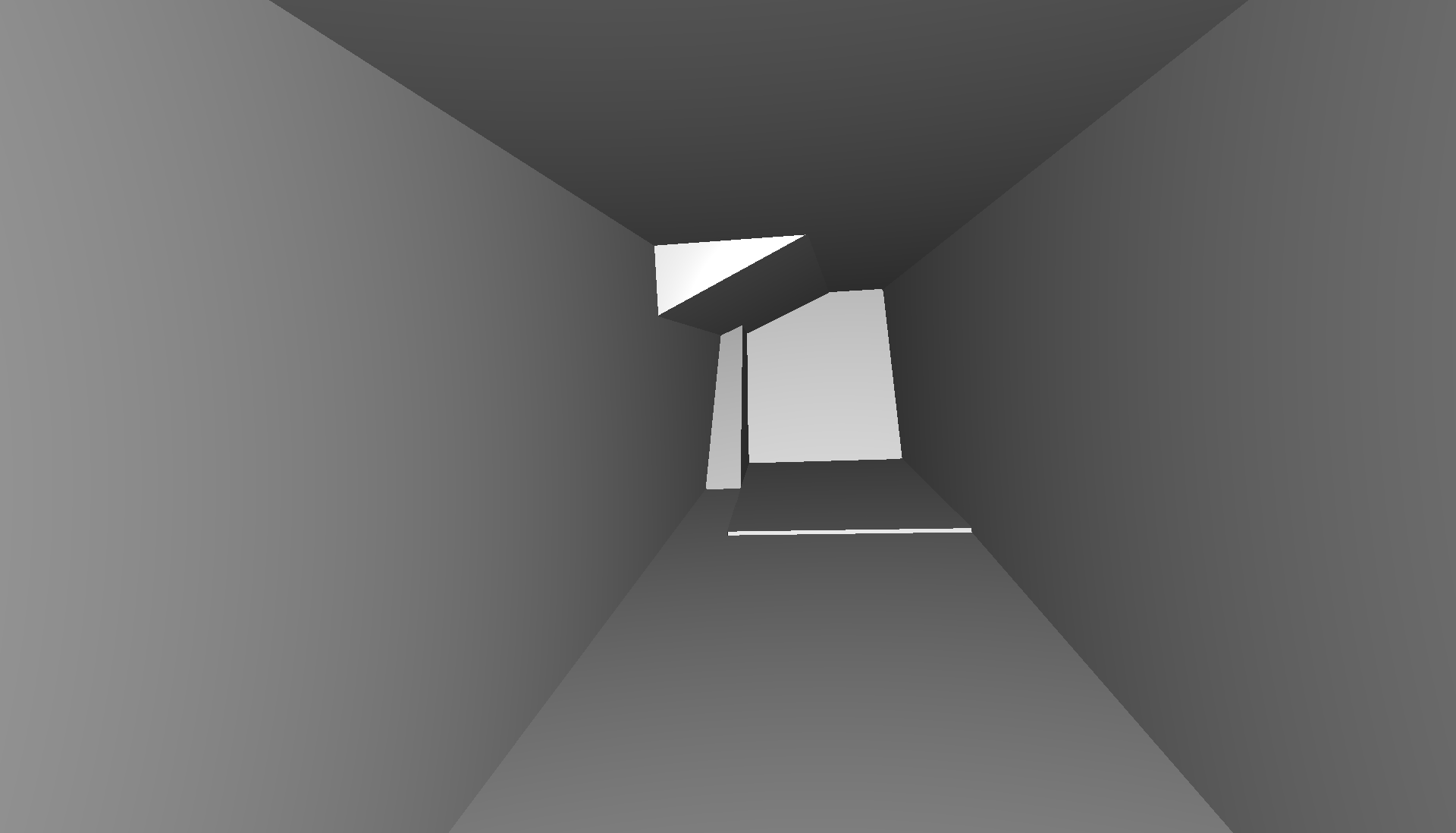}\raisebox{1mm}{\llap{(b6)\hspace*{20mm}}} &
\includegraphics[width=0.24\linewidth]{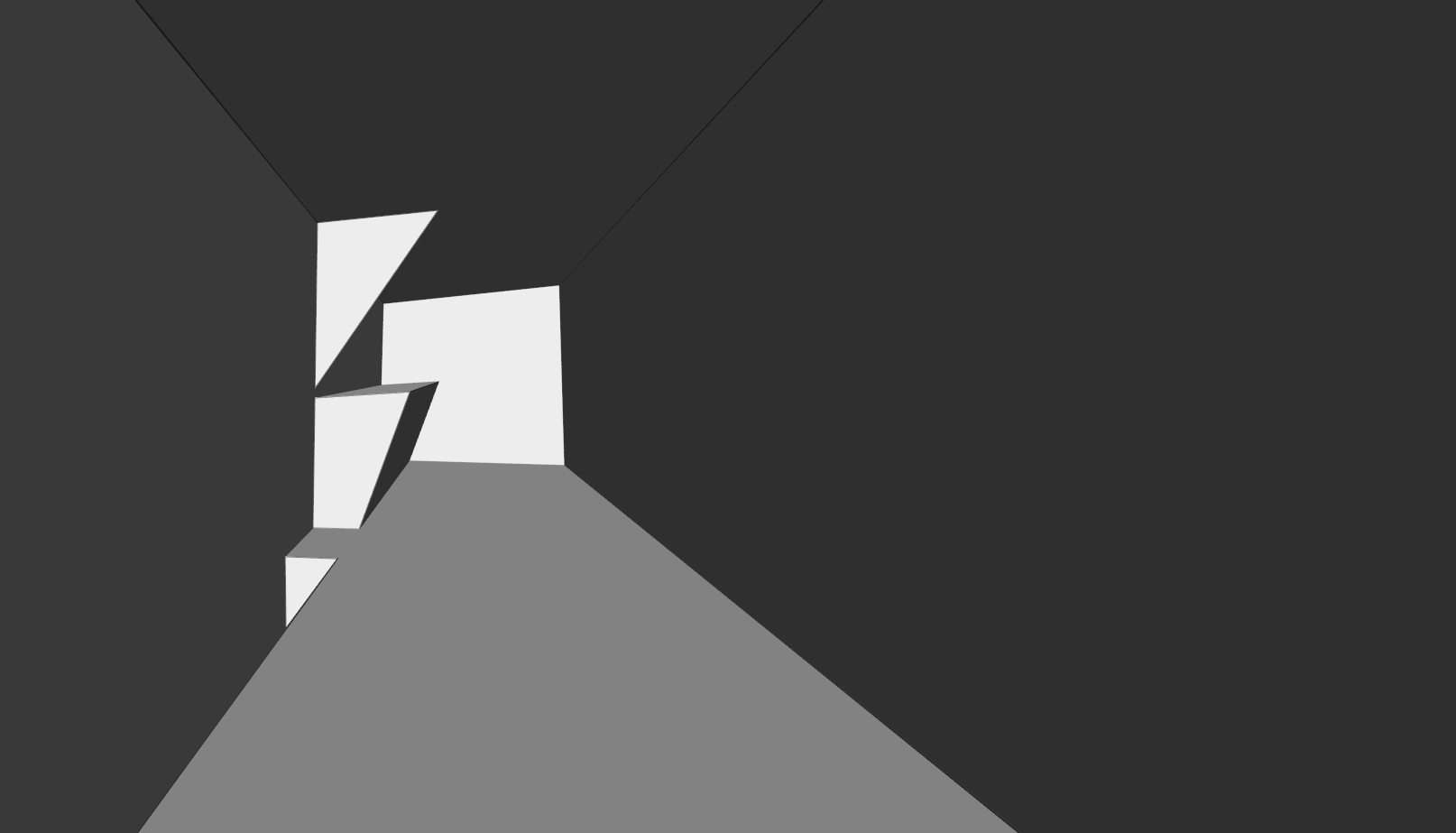}\raisebox{1mm}{\llap{(b7)\hspace*{20mm}}}
\end{tabular}
\vspace*{-2mm}
\caption{MeetingRoom (a), Terrains (b): (1) image sample, (2) segments from Line3D++ \cite{Line3Dpp,Hofer:cviu2016}, (3) our reconstruction, point-based reconstructions with Colmap \cite{SchoenbergerCVPR2016} then (4) Poisson~\cite{Kazhdan:sgp2006}, (5) Delaunay~\cite{labatut2009robust}, (6) Chauve et al.\cite{Chauve:cvpr2010}, (7) Polyfit \cite{NanICCV2017}.}
\vspace*{-4mm}
\label{fig:colmap}
\end{figure*}

Finally, we ran our plane detection and surface reconstruction, using a complete plane arrangement as baseline (see Sect.~\ref{sec:conclu}). Tab.~\ref{tab:def_params} lists default parameters for all datasets. We often had to tweak $\sigma_p$ of Line3D++ to get decent input lines, and sometimes our $\lambda_\iedge$~$=$ $\lambda_\icorner$ (see the supp.\,mat.\ for a sensitivity study).
Tab.~\ref{tab:stat} reports detection statistics. %

\paragraph*{Comparing to point-based reconstruction.}

To show the relevance of lines for scenes with little or no texture, in contrast to point-based methods (which are doubtlessly superior on textured scenes), 
we compare our method to a point-based piecewise-planar reconstruction~\cite{Chauve:cvpr2010} on HouseInterior
(cf.\ Fig.~\ref{fig:baseline}).
Even when densely sampling point on the ground-truth surface as seen from the viewpoints, \cite{Chauve:cvpr2010} yields a reconstruction with missing details (e.g., the lounge table) due to missing primitives in hidden area (e.g., under the table). Moreover, \cite{Chauve:cvpr2010} uses a regularization that minimizes the reconstructed area, which is relevant for points uniformly sampled on the surface but strongly penalizes unsampled regions (e.g., invisible planes of lounge table). In contrast, our method leads to a better plane discovery and a reconstruction robust to non-uniform sampling. (We also tried reconstructing from points sampled on the 3D lines, but the result is terrible; many planes are missed as points belong at most to one plane. As lines mostly lie on edges, the area cost also dominates the data term and creates holes in large planar regions.) More comparisons, also with Colmap \cite{SchoenbergerCVPR2016} and Polyfit \cite{NanICCV2017}, are on Fig.~\ref{fig:colmap}. The supp.\,mat.\ also studies the sensitivity to the number of images.

\paragraph*{Comparing to other line-based reconstruction methods.}

As said above, there are very few reconstruction methods based on lines. \cite{MentgesICRA2016} mostly reconstructs a soup of planes, sometimes with adjacencies, but without any topological guarantee. \cite{WittICRA2014} provides a slightly more behaved mesh, but reconstructions still look messy and overly simple, although usable enough for robotic planning. No code nor data are available for comparing with either of these methods.

\paragraph*{Quantitative evaluation.}

We evaluate the quality of reconstruction with two criteria: precision (proximity to the ground truth) and completeness (how much of the ground truth is reconstructed).
For this, we pick 2M points both on the reconstruction and on the ground truth, and we compute the nearest-neighbor distance from one set to the other.

Histograms of distances for \inhouse{} are plotted on Fig.~\ref{fig:baseline}(7).
Regarding precision, most of the points sampled on the reconstruction (91.4\%) lie at less than 5\,cm to the ground truth, showing that our RANSAC planes fit well the underlying surface and that our energy properly balances data fidelity and regularization.
The error profile for completeness is similar, and 95\% of the points on the ground truth are less than 8\,cm to the reconstruction.
It shows our regularization term do not over-smooth too much the surface by erasing details that would penalize completeness.

\paragraph*{Qualitative evaluation.}
Figs.~\ref{fig:datasets}, \ref{fig:baseline}, \ref{fig:colmap} illustrates our reconstructions with sparse data (Andalusian, DeliveryArea, MeetingRoom), and more (TimberFrame, Barn) or less texture (\inhouse{}), possibly with thin objects like beams (Bridge).
Compared to usual point clouds, our 3D line clouds are extremely sparse. Despite the noise on inliers and the number of outliers due to Line3D++, our method is able to reconstruct a good approximation of the scenes, which illustrates the robustness of our approach. Still Barn shows that it is hard to reconstruct a sieve-like shape (balcony) due to the visibility lines traversing it.

\paragraph*{Computation times.}

Although computing the visibility term is linear in the number of sub-segments, it is the most time-consuming part %
as it depends mostly on the number of cells in the plane arrangement, which is up to cubic in the number of %
planes. 
Time required for performing a whole reconstruction varies from 30 minutes (MeetingRoom) to 3 hours 30 minutes (TimberFrame).
Creating the linear program from scene data takes more time than solving it.

\begin{figure}[t!]
    \centering
    \vspace{-2mm}
    \includegraphics[width=\linewidth]{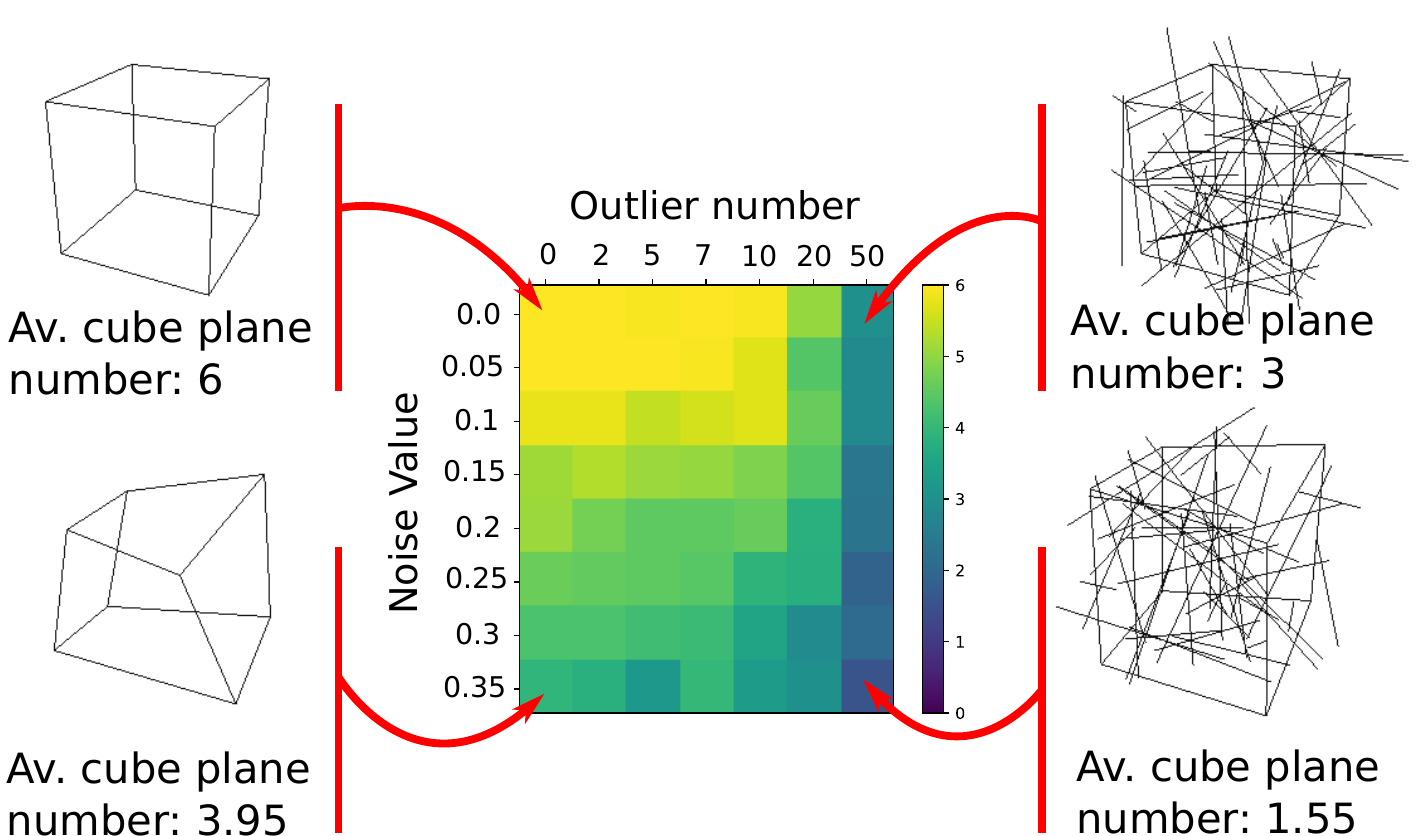}
    \vspace{-5mm}
    \caption{Robustness of RANSAC on lines for a cube defined by its edges. Value in grid is the average number of cube planes found, depending on the perturbation.}
    \label{fig:robust_noise_outliers}
    \vspace{-2mm}
\end{figure}

\paragraph*{Robustness of plane detection.}

To explore the robustness of our RANSAC formulation, we experimented with a toy example made of the 12 edges of a cube. We seek to extract the 6 planes associated to the 6 faces of the cube. We consider two types of perturbations: noise and outliers.

The cube has an edge length of~2.
We add noise to each segment endpoint, drawn from a uniform distribution with standard deviation ranging from 0 to~0.35.
Outliers, from 0 to~50, are segments generated by uniformly picking pairs of points in a 2-radius ball.
Finally for each couple (noise, \#outliers), we report the number of planes that include the 4 edges of an actual face of the cube, using parameters $\epsilon=\mbox{0.06}$ and $N_\iiter=\mbox{100}$, and averaging over 20 iterations.

Results are presented on Fig.~\ref{fig:robust_noise_outliers}. As expected, with a low level of perturbation, all planes are perfectly extracted. As the level of perturbation increases, for both noise and outliers, the rate of missed detections increases. Yet, even with a high level of noise, corresponding to a highly distorted cube (very non planar faces), we get a mean of 3.95 planes.

\paragraph*{Ablation study.} We compared with variants of RANSAC where (a) one line supports at most one plane, which leaves fewer lines for ulterior extractions and detects less planes, (b) we only consider $d(\lineseg,\plane) \leq \epsilon$ to decide if segment $\lineseg$ is an inlier to candidate plane $\plane$, ignoring if $\lineseg\,{\in}\,\Linesegs_1$, which misses many planes. We also tried ignoring the notion of structural lines at reconstruction time, treating segments with two supports as two ordinary lines (one for each plane),
which fails miserably. Last, we compared with a regularization using only corners or edges, which yields lower quality reconstructions. See supp.\,mat.\ for details and illustrations.

\section{Conclusion}
\label{sec:conclu}

We studied the specifics of line-based reconstruction and proposed the first method to create an intersection-free, watertight surface from observed line segments.  Experiments on synthetic and real data show it is robust to sparsity, outliers and noise, and that it outperforms point-based methods on datasets with little or no texture.

\paragraph*{Limitations and perspectives.}
The quality of 2D and 3D line segments at input (from Line3D++) is the main bottleneck of our method. Improving them would be very helpful.

Mainly, it would be specially relevant too to merge points and lines treatments into a single framework to offer a smooth transition from textured regions to textureless areas.

Also, in our experiments, we used the full-extent plane arrangement, i.e., with planes extending all the way to the scene bounding box. This is not intrinsic to our method; it merely provides a baseline. %
Because of a cubic complexity in the number of planes, the acceptable number of planes is limited to a few hundreds, which is in practice often enough for a single room or the exterior of a building, but not enough for a complete BIM model.  (It is also easy to keep the best few hundred planes after RANSAC detection to make sure the pipeline succeeds.)  Yet, preliminary experiments with a coarse-to-fine approach show promising results for scaling to large scenes. In the cell complex, limiting the plane extent with a heuristic on a coarse voxel-based partition \cite{Chauve:cvpr2010} or adapting 2D kinetic polygonal plane partitioning \cite{BauchetCVPR2018} to 3D would also reduce the complexity.
Moreover, defining a notion of extent for line-detected planes, similar to $\alpha$-shapes in the case of points \cite{EdelsbrunnerTIT1983} but adapted to lines \cite{KreveldCGF2011,KreveldCGF2013}, could also be used to introduce so-called `ghost planes', corresponding to unobserved, hidden planes at occluding edges of observed surfaces \cite{Chauve:cvpr2010,Boulch:sgp2014}.

Last, global regularization weights favor highly sampled surfaces. Adapting them to be more sensitive to weakly supported surfaces as in~\cite{jancosek2011multi} could improve the results.

{\small
\bibliographystyle{ieee}
\bibliography{egbib}
}

\end{document}